\documentclass[preprint, 5p, times, twocolumn]{elsarticle}

\usepackage{amssymb}
\usepackage{graphicx}
\usepackage{amsmath} 
\usepackage{amsthm}
\usepackage{float}
\usepackage{booktabs}
\usepackage[english]{babel}
\usepackage[ruled]{algorithm2e}
\usepackage{comment}
\usepackage{mathrsfs}
\usepackage{hyperref}
\usepackage{tabularx}
\usepackage{xcolor}
\usepackage{multirow}
\usepackage{pifont}
\usepackage{subcaption}
\usepackage{tikz}
\usepackage{wrapfig}

\newcommand{\hh}[1]{\textcolor{black}{#1}}

\journal{Robotics and Autonomous Systems}

\newcommand\copyrighttext{%
  \footnotesize \textcopyright This paper has been accepted for publication at Robotics and Autonomous Systems. Please, when citing the paper, refer to the official manuscript with the following DOI: 10.1016/j.robot.2025.105020.}
\newcommand\copyrightnotice{%
\begin{tikzpicture}[remember picture,overlay]
\node[anchor=south,yshift=10pt] at (current page.south) {\fbox{\parbox{\dimexpr\textwidth-\fboxsep-\fboxrule\relax}{\copyrighttext}}};
\end{tikzpicture}%
}

\begin{document}

\theoremstyle{definition}
\newtheorem{definition}{Definition}

\theoremstyle{remark}
\newtheorem{remark}{Remark}

\begin{frontmatter}



\title{RUMOR: Reinforcement learning for Understanding a Model of the Real World for Navigation in Dynamic Environments}


\author[label1]{Diego Martinez-Baselga\corref{cor1}}
\ead{diegomartinez@unizar.es}
\author[label1]{Luis Riazuelo}
\ead{riazuelo@unizar.es}
\author[label1]{Luis Montano}
\ead{montano@unizar.es}
\cortext[cor1]{Corresponding author at: Robotics, Perception and Real Time Group, Aragon Institute of Engineering Research (I3A), University of Zaragoza, Spain.}
\affiliation[label1]{organization={Robotics, Perception and Real Time Group, Aragon Institute of Engineering Research (I3A), University of Zaragoza},
            city={Zaragoza},
            postcode={50018}, 
            country={Spain}}

\begin{abstract}
Autonomous navigation in dynamic environments is a complex but essential task for autonomous robots, with recent deep reinforcement learning approaches showing promising results. However, the complexity of the real world makes it infeasible to train agents in every possible scenario configuration. Moreover, existing methods typically overlook factors such as robot kinodynamic constraints, or assume perfect knowledge of the environment. In this work, we present RUMOR, a novel planner for differential-drive robots that uses deep reinforcement learning to navigate in highly dynamic environments. Unlike other end-to-end DRL planners, it uses a descriptive robocentric velocity space model to extract the dynamic environment information, enhancing training effectiveness and scenario interpretation. Additionally, we propose an action space that inherently considers robot kinodynamics and train it in a simulator that reproduces the real world problematic aspects, reducing the gap between the reality and simulation. We extensively compare RUMOR with other state-of-the-art approaches, demonstrating a better performance, and provide a detailed analysis of the results. Finally, we validate RUMOR's performance in real-world settings by deploying it on a ground robot. Our experiments, conducted in crowded scenarios and unseen environments, confirm the algorithm's robustness and transferability.
\end{abstract}


\begin{highlights}
\item A remarkable performance in robot navigation in dynamic real-world scenarios.
\item Deep reinforcement learning (DRL) proves to be effective at decision making.
\item The DOVS model can extract the scenario dynamism and its future information.
\item The DOVS top of DRL improves its efficiency and generalization.
\item RUMOR already considers differential drive kinodynamics in the DRL action space.
\end{highlights}


\end{frontmatter}

\copyrightnotice



\section{Introduction}

Motion planning and navigation in dynamic scenarios is essential for autonomous robots, for applications as delivery or assistance. Nevertheless, traditional planners fail in environments where the map is mutable or obstacles are dynamic, leading to suboptimal trajectories or collisions. Those planners typically consider only the current obstacles' position measured by the sensors, without considering the future trajectories they may have.

New approaches that try to solve this issue include promising end-to-end Deep Reinforcement Learning (DRL) based methods. The robot learns a policy that selects the best action for each of the situations, directly represented with the sensed information. They present results that outperform model-based approaches in terms of success (reaching the goal without collisions) and time to reach the goal. However, the complexity of the real world and the huge variety of different possible situations the robot may encounter, with different number of obstacles, different shapes and different behaviors, poses a huge training challenge for these end-to-end approaches. Moreover, they typically lack in sim2real transfer capabilities, not considering robot kinodynamic constraints, or partial observability issues.

\begin{figure}
    \centering
    \includegraphics[width=0.61\linewidth,trim={7cm 1.cm 7cm 1.0cm},clip]{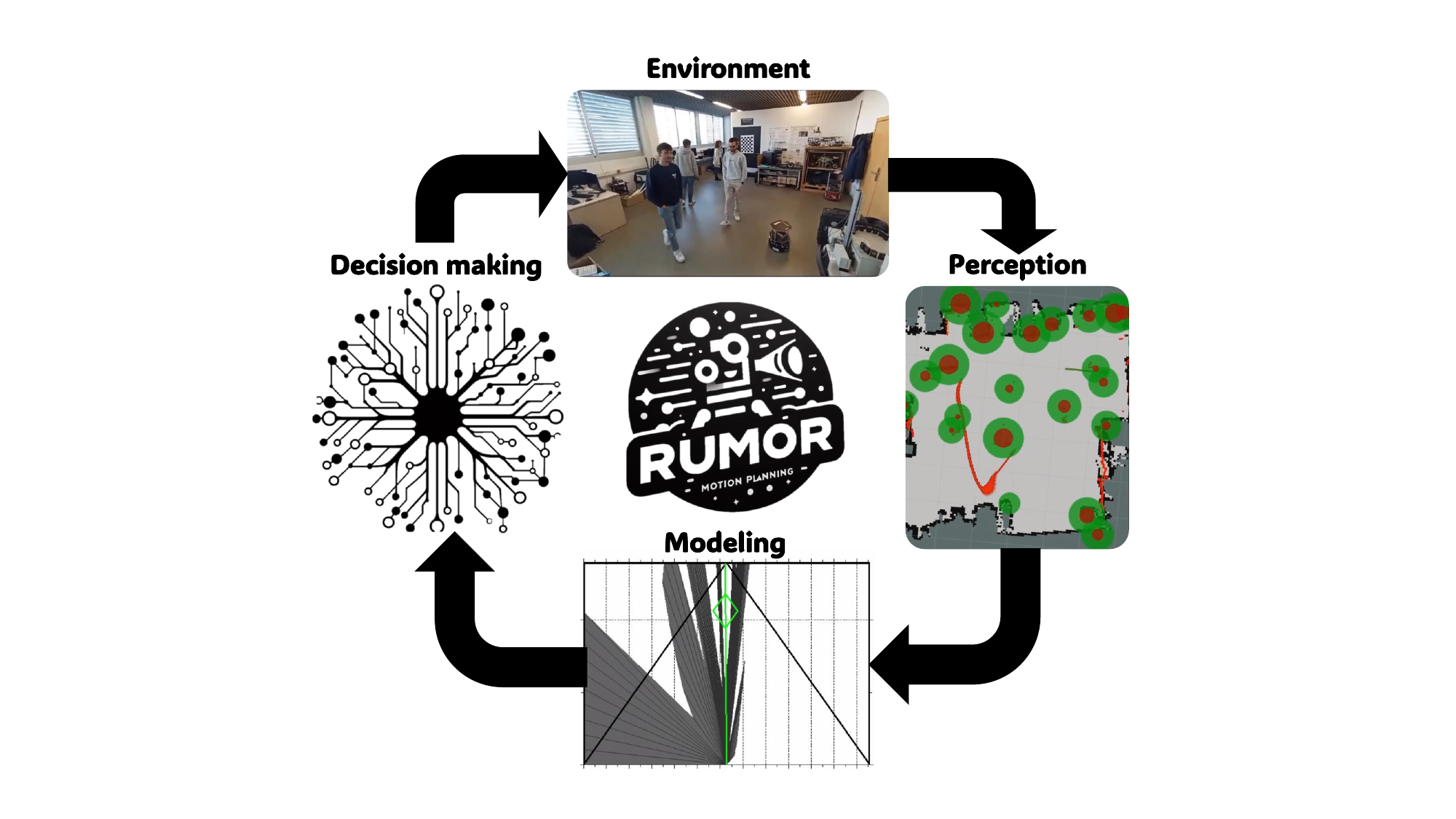}
    \caption{Pipeline of the approach presented. It takes the information sensed from the environment to construct a model of the dynamism of the scenario. Then DRL is used to compute differential-drive velocity commands.}
    \label{fig:main-fig}
\end{figure}

In this paper, we propose \textbf{RUMOR} (\textbf{R}einforcement learning for \textbf{U}nderstanding a \textbf{MO}del of the \textbf{R}eal world), a motion planner that combines model-based and DRL benefits. We use the deep abstraction of the environment provided by the Dynamic Object Velocity Space (DOVS) \cite{lorente2018model} to understand the surrounding scenario. The DOVS model represents the dynamism and the future of the environment in the velocity space of the robot. The planner applies this information as an input of the DRL algorithm, learning to interpret the future dynamic scenario knowledge, taking advantage over other approaches that use raw obstacle information and are not able to generalize in scenarios that are different from those previously experienced. The perception is decoupled from the learning algorithm and used to construct the DOVS, making it able to work with any sensor as a LiDAR or a camera. Figure~\ref{fig:main-fig} provides a representation of RUMOR pipeline. In addition, unlike other methods that simply choose a velocity ranging from the maximum and minimum robot velocities, we propose an action space that inherently considers differential-drive kinodynamic restrictions, improving the applicability of RUMOR in real robots. The algorithm is trained and tested in a realistic simulator, where all information is extracted from the sensor measurements, even the own robot localization. Training using the robot constraints and a simulator that includes the problems of the real world reduces the sim2real transfer complexity. 

In Section~\ref{sec:background}, we analyze the background and relate our work with the state of the art and in Section~\ref{sec:preliminaries} we state the preliminaries needed for understanding the method. Section~\ref{sec:DRL-setup} and Section~\ref{sec:network} present the approach taken. We provide comparisons of RUMOR with other method of the state-of-the-art and experiments with a real-ground robot in Section~\ref{sec:results}. Finally, the conclusions of the work are stated in Section~\ref{sec:conclusion}. 

Our contributions are summarized in:

\begin{itemize}
    \item A motion planner that combines a very complete and descriptive information of the environment on top of DRL to merge the benefits of model-based and DRL methods.
    \item A way to inherently consider differential-drive robot kinodynamics in a DRL planner and deal with some real-world problems as the variety of scenario configurations or partial observability.
    \item Experiments that show the benefits of our contribution, comparisons with other methods and real-world demonstrations.
\end{itemize}

The code and videos are available in \url{https://github.com/dmartinezbaselga/RUMOR}.
\section{\hh{Related work}}\label{sec:background}

\hh{In this section, we give a brief overview of the existing solutions for autonomous navigation in dynamic environments, with special emphasis in deep reinforcement learning planners. We refer the readers to recent surveys~\cite{mavrogiannis2023core,francis2023principles,singamaneni2024survey} for a more detailed introduction to the field.}

\subsection{Motion planning in dynamic environments}\label{sec:}

\hh{The motion planning problem is frequently solved with the conjunction of a global planner, which computes a static path considering a continuous and static map, and a local planner that follows it accounting for the dynamics of the robot and obstacles that were not considered by the global planner. Traditional local planners like the Dynamic Window Approach (DWA)~\cite{fox1997dynamic} do not consider that obstacles may move in time, thus leading} to collisions and suboptimal trajectories, as seen in some works as the benchmark proposed in \cite{kastner2022arena}. \hh{First} approaches for dynamic environments \hh{use attractive forces to lead the robot to its goal and repulsive forces to drive it away from obstacles}, such as artificial potential fields \cite{qixin2006evolutionary} or \hh{Social Force} \cite{helbing1995social}. They were later improved with Time Elastic Bands (TEB) \cite{rosmann2015timed} or multiple extensions to Social Force \cite{ferrer2013robot, jiang2017extended}.

A big group of works are velocity-space based. The \textit{Velocity Obstacle (VO)}, introduced in  \cite{fiorini1998motion}, refers to the set of \hh{possible robot} velocities that could lead \hh{it} to a collision with an obstacle that has a certain velocity in the near future. Based on the VO concept,  the \textit{Dynamic Object Velocity Space (DOVS)} \hh{for differential-drive robots }is defined in \cite{lorente2018model} as the robot velocity space that includes \hh{the safe and unsafe robot possible velocities for a time horizon derived from VO}. In that work, a planner based on strategies is also defined, the S-DOVS. Another work \cite{mackay2022rl} uses the DOVS to discard velocities that lead to collisions from the action space of a tables-based reinforcement learning algorithm. However, unlike our work, it does not use the information of the DOVS to interpret the environment and just filters the unsafe actions available. In addition, the results presented in their work are very similar to S-DOVS, which does not require training time. Other extensions of VO use reciprocal collision assumptions, such as Reciprocal Velocity Obstacles (RVO) \cite{van2008reciprocal} or Optimal Reciprocal Collision Avoidance (ORCA) \cite{van2011reciprocal}, which leads to failures in scenarios where there are non-collaborative obstacles.

Application of optimization-based approaches in dynamic environments, such as those that are based in a Model Predictive Controller (MPC), is gaining ongoing attention in recent years. The authors of \cite{brito2019model} present a Model Predictive Contouring Controller that considers the future obstacle trajectories to safely reach the goal. It was recently combined with a global planning approach \cite{de2023globally} to dynamically change between possible global plans. Other approaches combine crowd motion prediction with MPC navigation \cite{10341464} or introduce topological invariance to improve social navigation \cite{mavrogiannis2022winding}. As our approach, these works account for kinematic and dynamic constraints of the robot, but they have the problem of being computationally costly, and their planning is reliable as long as the predictions are accurate, so they are not suitable for very dense or chaotic scenarios.

\subsection{Deep reinforcement learning planners}

Deep Reinforcement Learning (DRL) \cite{mnih2015human} is a method used to learn to estimate the optimal policy that optimizes the cumulative reward obtained in an episode with a deep neural network. \hh{DRL algorithms have proven state-of-the-art performance in several benchmarks}, including Rainbow \cite{hessel2018rainbow} for discrete action spaces, distributed reinforcement learning algorithms \cite{horgan2018distributed} or actor-critic methods \cite{haarnoja2018soft}. 

Some works \hh{study the benefits} of DRL in robot motion planning and the limitations of traditional planners in dynamic environments. Defining strategies for every situation that may be found in the real world or optimizing trajectories in very dynamic environments is intractable, so DRL \hh{emerges to} efficiently solve the decision-making problem, which is complex and has many degrees of freedom. There are works \cite{kastner2022arena,kastner2023arena,Kästner-RSS-24} that implement and compare DRL algorithms with model-based ones, proving their performance. 

\hh{There are two big groups of DRL planners: end-to-end and agent-based planners. The former use the raw sensor measurements directly as the DRL network input. These measurements may be sensor maps \cite{yao2021crowd,qiu2022learning,yu2024pathrl,yu2024ldp}, LiDAR laser scans~\cite{pfeiffer2017perception,guldenring2020learning,xie2023drl, damanik2024lics} or images~\cite{dugas2022navdreams,zhao2023learning,song2023learning} taken from robot's onboard sensors. In spite of the efforts in designing realistic simulators~\cite{tsoi2022sean, Kästner-RSS-24, puighabitat}, these methods still experience distribution-shift problems when the deployment environments differ from the training one~\cite{TriessDreissigRist2021_1000157575, 10611710}, and require the robot to have the same sensor configuration (for example, sensor position, range or resolution) both in training and in deployment, limiting the applicability of their policy. Agent-based planners extract an agent-level representation from sensor measurements and use it as the planner input. In this way, the robot does not require to have specific sensor settings and does not experience distribution shifts due to the scenario geometry and shape. Furthermore, it may learn to understand deeper agent-level behaviors by differentiating decision-making agents from other obstacles~\cite{everett2018motion}.}

The first agent-based DRL approach proposed for motion planning \cite{chen2017decentralized} uses a fully connected network to select velocities regarding the position and velocity of the closest surrounding obstacles. LSTM \cite{hochreiter1997long} layers were included in \cite{everett2018motion,everett2021collision} to account for multiple obstacles. SARL \cite{chen2019crowd} uses attention to model interactions among obstacles, achieving state-of-the-art performance. Recent approaches compare their results to it, and they usually focus on improving the structure of the network. For example, \cite{chen2020relational} uses a graph neural network or \cite{liu2023intention} an attention-based interaction graph. \hh{In a recent study, } \cite{10160876} \hh{we showed that the performance of DRL approaches may be improved using intrinsic rewards}. The use of DRL algorithms in robot navigation has many important challenges, including the need of learning from a limited subset of possible scenarios, environmental and robot constraints or partial observability \cite{ibarz2021train,dulac2021challenges}. The previous approaches do not face these problems. In our approach, instead of directly process the observations, we use the DOVS model, that represents the dynamism of the environment in the velocity space, as an input to the DRL algorithm, doing the decision-making process agnostic of the specific scenario.

Other works consider social settings, such as \cite{hu2022crowd} including social stress indices, \cite{zhou2022robot} a social attention mechanism or \cite{xue2023crowd} social rewards to shape the robot behavior. Another approach \cite{yang2023rmrl} uses a social risk map as part of the input of the network. The applicability of these social approaches is limited when other obstacles are not humans, like other robots, or humans do not behave as expected.

 An example of an approach that considers the kinodynamic restrictions is \cite{patel2021dwa}, which combines DRL with DWA \cite{fox1997dynamic}, but only achieving a success rate of 0.54 in low dynamic scenarios. Our proposed approach solves the problem of selecting feasible velocities by considering kinematic and dynamic equations in the action space formulation, limiting the downgrade in performance.
\section{\hh{Problem formulation}}\label{sec:preliminaries}

We consider a \hh{differential-drive} robot \hh{that shares the workspace $\mathcal{W} \subseteq \mathbb{R}^2$ with static and dynamic obstacles. The robot} position at time $t$ is \hh{$\mathbf{p_t}=(x_t,y_t,\theta_t) \in \mathcal{W}\times[-\pi, \pi)$} and its velocity \hh{$\mathbf{u}_t = (\omega_t, v_t) \in \mathcal{V} \subseteq \mathbb{R}^2$} (angular and linear velocity). \hh{The robot maximum linear and angular velocities are bounded: $\mathcal{V}=[-\omega_{max}, \omega_{max}]\times[0,v_{max}]$.} It has to reach a goal \hh{$\mathbf{g} = (x_g, y_g) \in \mathcal{W}$} in the minimum possible time, while avoiding static and dynamic obstacles in the environment. The robot is represented with a disk with radius $r_{robot}$, and the dynamic obstacles are assumed to be circular disks with radius $\mathbf{r}_{obs} \in \mathbb{R}^M$, where $M$ is the number of dynamic obstacles. \hh{For each obstacle $i$, we assume that the robot may estimate its current position and velocity using the sensor information. }Let \hh{$\mathcal{R}_t(\mathbf{u})\in\mathcal{W}\times[-\pi, \pi)$} be the robot state at time $t$ \hh{if it drives with} velocity $\mathbf{u}$, \hh{$\mathscr{O}_{i,t} \in \mathcal{W}$ the set of points occupied by obstacle $i$ at time $t$, $\mathscr{O}_{i,t}^{T_h}=[\mathscr{O}_{i,t},\dots, \mathscr{O}_{i,t+T_h}]$ for the time horizon $T_h$ and $\mathscr{O}_{\forall,t}^{T_h} = [\mathscr{O}_{1,t}^{T_h}, \dots, \mathscr{O}_{M,t}^{T_h}]$ for the whole set of static and dynamic obstacles in the environment.} 
\hh{We consider that the robot follows the unicycle model:}
\begin{equation}
    \begin{pmatrix} \dot{x}_t \\ \dot{y}_t \\ \dot{\theta}_t
    \end{pmatrix} = 
    \begin{pmatrix} \cos{\theta_t} & 0 \\ \sin{\theta_t} & 0 \\ 0 & 1
    \end{pmatrix} \begin{pmatrix} v_t \\ \omega_t
    \end{pmatrix}
    \label{eq:kinematics}
\end{equation}

\hh{The robot is also restricted by }the low-level forward model that relates the wheel velocities to the actions $\mathbf{u}_t$ for a differential-drive robot:
\begin{equation}
\label{eq:wheels}
\begin{split}
    & v_t = \frac{r_w(\omega_{r,t} +\omega_{l,t})}{2} \\
    & \omega_t = \frac{r_w(\omega_{r,t} -\omega_{l,t})}{L_w} \\
\end{split}
\end{equation}
\noindent where $r_w$ is the wheel radius, $\omega_{r,t}$ and $\omega_{l,t}$ the angular velocity of the right and left wheels, respectively, and $L_w$ the distance between wheels. \hh{The relationship $\frac{v_t}{\omega_t}=\frac{2}{L_w}$ imposes two linear constraints regarding} the maximum \hh{linear velocity} a robot may take in a particular moment \hh{with respect to its angular velocity}:

\begin{equation}
    \begin{split}
            & l_1: v_t \leq \frac{v_{max}}{\omega_{max}}\omega_t + v_{max}, \quad \omega_t \leq 0 \\
            & l_2: v_t \leq -\frac{v_{max}}{\omega_{max}}\omega_t + v_{max}, \quad \omega_t > 0 \\
    \end{split}
    \label{eq:dymanic-const}
\end{equation}

\hh{We consider that the} robot has a maximum linear acceleration of $a_{max}$. The change in the linear velocity of the robot is limited by its previous linear velocity and, due to Equation~\ref{eq:wheels}, its velocity limits:

\begin{equation}
    \begin{split}
        & v_{t+1} \leq \left\lbrace
        \begin{aligned}
            \frac{v_{max}}{\omega_{max}}\omega_t + v_t + a_{max} \Delta t, & \quad \omega_t \leq 0 \\
            -\frac{v_{max}}{\omega_{max}}\omega_t + v_t + a_{max} \Delta t, & \quad \omega_t > 0 \\
        \end{aligned}
        \right. \\
        & v_{t+1} \geq \left\lbrace
        \begin{aligned}
            -\frac{v_{max}}{\omega_{max}}\omega_t + v_t - a_{max} \Delta t, & \quad \omega_t \leq 0 \\
            \frac{v_{max}}{\omega_{max}}\omega_t + v_t - a_{max} \Delta t, & \quad \omega_t > 0 \\
        \end{aligned}
        \right. \\
    \end{split}
    \label{eq:acc-const}
\end{equation}

\begin{figure}
    \centering
    \includegraphics[width=0.47\textwidth]{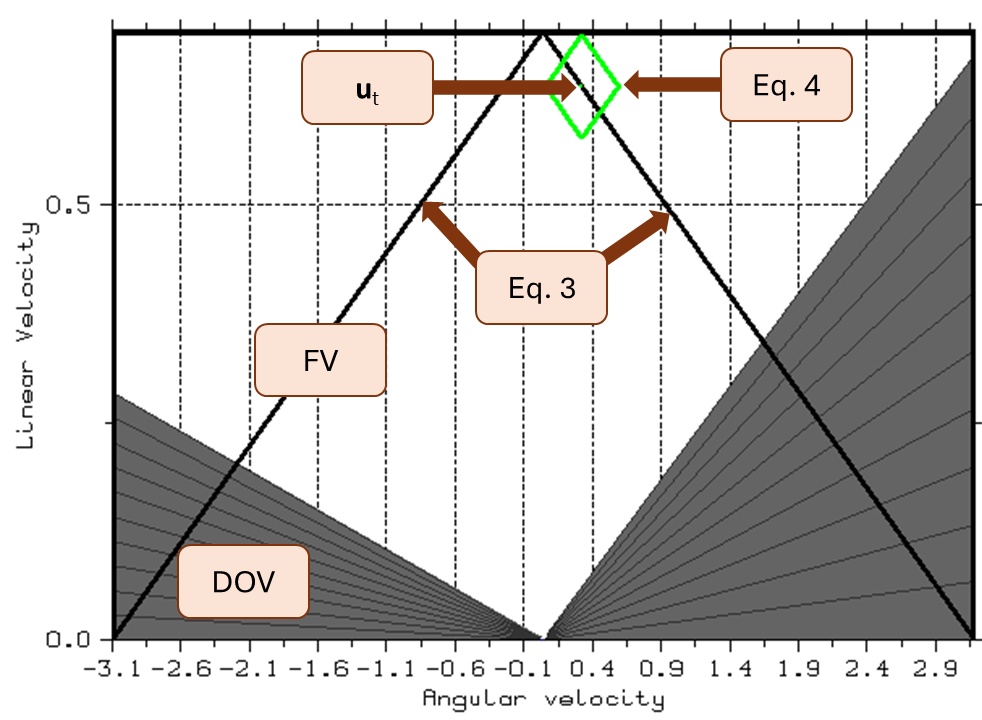}
    \caption{Graphical representation of the DOVS model \hh{and the differential-drive robot restrictions. Constraints of Equation~\ref{eq:dymanic-const} are represented with two black lines and restrict maximum linear velocities regarding the angular velocity. Equation~\ref{eq:acc-const} are plotted as a green rhombus around $\boldsymbol{u}_t$, representing acceleration limits regarding a differential-drive robot.} In this example, the robot velocity limits are $v_{max}=0.7$ m/s, $\omega_{max}=\pi$ rad/s and $a_{max}=0.3$ m/s². \hh{The dark (DOV) and white (FV) areas include unsafe and safe velocities derived using VO for a time horizon.}}
    \label{fig:DOVS-rhombus}
\end{figure}

The restrictions of Equation~\ref{eq:dymanic-const} and Equation~\ref{eq:acc-const} are plotted in a graphical representation of the robot velocity space in Figure~\ref{fig:DOVS-rhombus}. \hh{The velocity in the next control period $\boldsymbol{u}_{t+1}$ can never be above the two big black lines that relate linear and angular velocity constraints, and can never be outside the green dynamic window that surround $\boldsymbol{u}_t$, which constraints linear and angular acceleration and their relationship.}

\section{Methodology}

In this section, we explain the concepts of the DOVS and its adaptation used in RUMOR, as well as the deep reinforcement learning solution proposed to solve the problem.

\subsection{Dynamic Object Velocity Space (DOVS)}\label{sec:dovs}

In this work, we use the DOVS~\cite{lorente2018model} to model the information of the environment. It translates scenario information from the workspace $\mathcal{W}$ to the robot velocity space $\mathcal{V}$, by computing the velocities in $\mathcal{V}$ lead to a collision within a temporal horizon. This provides a robocentric, bounded, scalable and flexible model with the original information of the workspace.

The DOVS is modeled using possible robot velocities. As we consider a differential-drive robot, its velocities result in circular trajectories if they are kept constant. We denote $\tau_j \equiv \tau_j(\omega_j,v_j)$ the robot trajectory $\tau_j$ produced by $\boldsymbol{u}_j=(\omega_j, v_j)\in\mathcal{V}$ and $\mathcal{T}$ the collection of different circular trajectories that collectively cover $\mathcal{V}$. The set of colliding velocities with the obstacle $i$ are the trajectories that intersect with $i$ within a time horizon. They define Dynamic Object Velocity of $i$:

\begin{definition}
The Dynamic Object Velocity (DOV) of obstacle $i$, for the set of trajectories $\mathcal{T}$ and the time horizon $T_h$ is defined as:
\begin{equation}
\begin{split}
    DOV(\mathscr{O}_{i, t}^{T_h}, \mathcal{T}) = & \left\{\mathbf{u}_j=(\omega_j, v_j) \in \mathcal{V} \mid \exists \tau_j \in \mathcal{T},\; \right.\\
    & \left. \exists t' \in [t,t+T_h],\; \mathcal{R}_{t'}(\mathbf{u}_j)\cap\mathscr{O}_{i,t'} \neq \emptyset \right\}\\
\end{split}
\end{equation}
\end{definition}

The DOV of all obstacles in the environment is combined for the general definition:
\begin{definition}
The Dynamic Object Velocity (DOV) of $M$ obstacles of the set of trajectories $\mathcal{T}$ and the time horizon $T_h$ is defined as:
\begin{equation}
    DOV(\mathscr{O}_{\forall,t}^{T_h}, \mathcal{T}) = \bigcup_{i=1}^M DOV(\mathscr{O}_{i, t}^{T_h}, \mathcal{T})
\end{equation}
\end{definition}

The Free Velocity (FV) is constructed as the complementary set of the DOV with the collision-free velocities:

\begin{definition}
The Free Velocity (FV), for the set of trajectories $\mathcal{T}$ and the time horizon $T_h$ is defined as:
\begin{equation}
\begin{split}
    FV(\mathscr{O}_{\forall, t}^{T_h}, \mathcal{T}) = & \left\{\mathbf{u}_j=(\omega_j, v_j) \in \mathcal{V} \mid \mathbf{u}_j \notin DOV(\mathscr{O}_{\forall, t}, \mathcal{T}) \right\}\\
\end{split}
\end{equation}
\end{definition}

Finally, the DOVS is modeled as the union of the whole set of velocities that comprise the DOV and the FV, categorizing every velocity in $\mathcal{V}$:

\begin{definition}
The Dynamic Object Velocity Space (DOVS), for the set of trajectories $\mathcal{T}$ and the time horizon $T_h$ is defined as:
\begin{equation}
    DOVS(\mathscr{O}_{\forall, t}^{T_h}, \mathcal{T}) = \left\{DOV(\mathscr{O}_{\forall, t}^{T_h}, \mathcal{T}) \cup FV(\mathscr{O}_{\forall, t}^{T_h}, \mathcal{T})\right\}
\end{equation}
\end{definition}

\begin{figure}
    \centering
    \begin{tabular}{@{}cc@{}}
            \includegraphics[width=0.47\linewidth,trim={0.1cm 0 31.cm 0},clip]{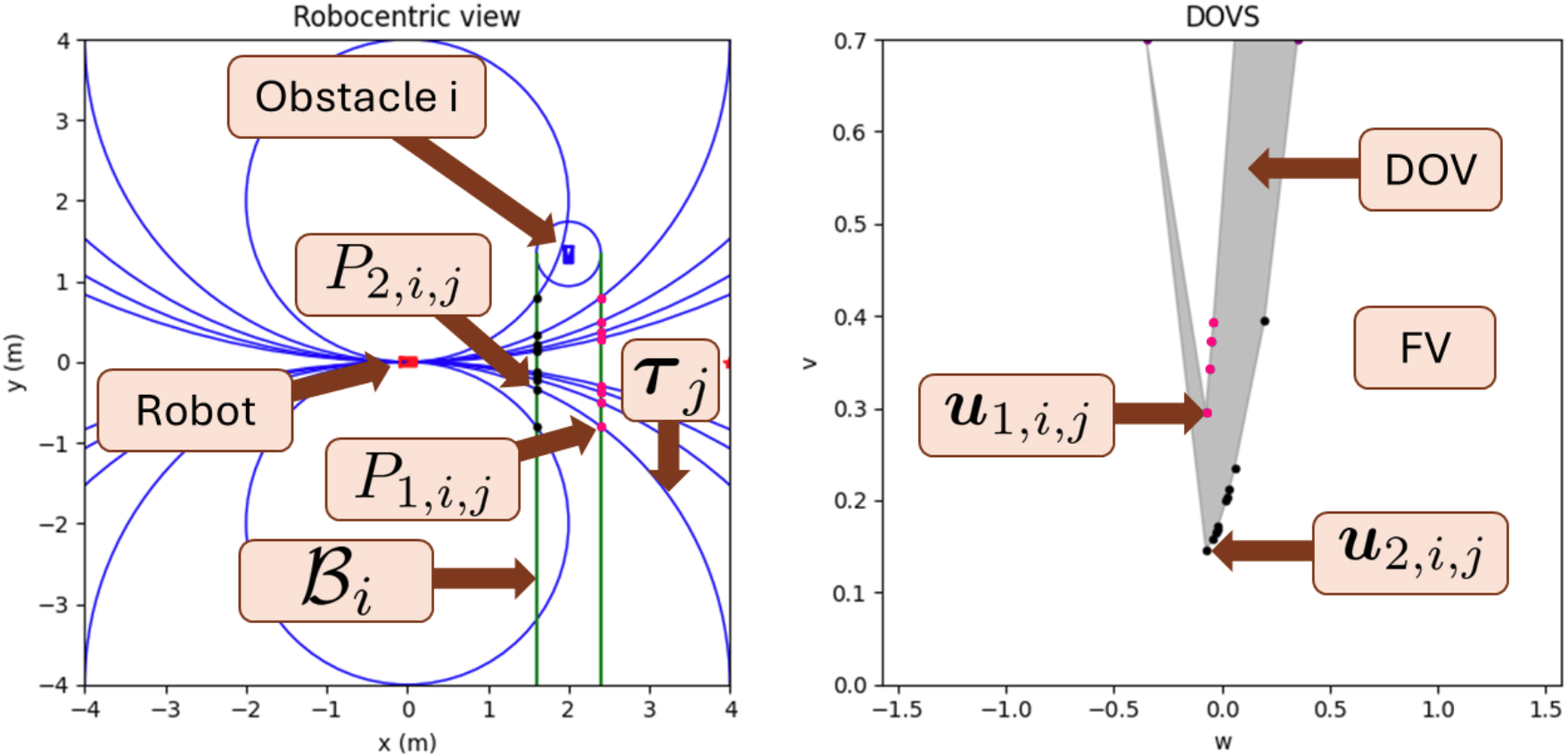} &
            \includegraphics[width=0.47\linewidth, trim={31.cm 0 0.1cm 0},clip]{DOVS_ws.png}\\
        (a) & (b)
    \end{tabular}
    \caption{A robocentric view of $\mathcal{W}$ (a) and a graphical representation of the DOVS (b) of a scenario where a robot faces the obstacle $i$ that follows a linear trajectory. In (a), the robot center is represented in red, the trajectories \hh{$\tau_j\in\mathcal{T}$} sampled in blue lines, the obstacle augmented radius with a blue circle and the collision band $\mathcal{B}_i$ with green lines. The intersection points between $\tau_j$ and $\mathcal{B}_i$ are $P_{1,i,j}$ (right) and $P_{2,i,j}$ (left). In \hh{(b)}, the maximum velocities to pass after the obstacle are represented with black dots ($\boldsymbol{u}_{2,i,j}$) and the minimum velocities to pass before it with purple dots ($\boldsymbol{u}_{1,i,j}$). The \hh{DOV} is represented in gray \hh{and the FV in white}.}
    \label{fig:workspace}
\end{figure}

The DOVS is computed with the concept of the collision band, $\mathcal{B}_i$. The robot is reduced to a point and the obstacle is enlarged with the robot radius, to simplify computations while keeping the same collision area. The collision band, $\mathcal{B}_i$, is the area that obstacle $i$ will occupy by following its predicted trajectory. In this work, we assume the robot only knows obstacle $i$ current velocity, so its trajectory is predicted by considering that its velocity is kept constant. Thus, obstacle trajectories are straight lines or circular trajectories, depending on whether the obstacle has angular velocity or not. The DOVS is computed with the intersections of trajectories and $\mathcal{B}_i$. For each robot circular trajectory $\tau_j \in \mathcal{T}$, the intersection points between the two lines limiting $\mathcal{B}_i$ and $\tau_j$ are $P_{1,i,j}(x_{1,i,j}, y_{1,i,j})$ and $P_{2,i,j}(x_{2,i,j}, y_{2,i,j})$. They are graphically represented in Figure~\ref{fig:workspace} (a) in a robocentric view for a few example trajectories. 

\hh{Different combinations of linear and angular velocities lead to the same trajectory, as long as the trajectory radius, $\frac{v}{\omega}$, is the same. Depending on the velocity, the robot will eventually be at the same time as obstacle $i$ in $\mathcal{B}_i$, colliding, or will not.} The time $t_{2,i,j}$, which is the time any point of the obstacle $i$ takes to reach $P_{2,i,j}$, and $t_{1,i,j}$, which is the time it takes to stop being in contact to $P_{1,i,j}$\hh{, may be efficiently computed considering the velocity of $i$~\cite{lorente2018model}. We denote $\boldsymbol{u}_{k,i,j}$ the velocity that makes the robot follow trajectory $\boldsymbol{\tau}_j$ and reach $P_{k,i,j}$ at time $t_{k,i,j}$, for $k=1,2$. Colliding velocities are velocities $\boldsymbol{u}_j$ that follow $\boldsymbol{\tau}_j$ and $v_{2,i,j}\leq v_j \leq v_{1,i,j}$. Therefore, $v_{1,i,j}$ is the robot minimum linear velocity to pass before the obstacle passes without collision, and $v_{2,i,j}$ the maximum linear velocity to pass after it. They are represented in Figure~\ref{fig:workspace}(b), and} they are computed as:

\begin{equation}
\begin{split}
      & \omega_{k,i,j}=\frac{\theta_{k,j}}{t_{k,i,j}} = \frac{\operatorname{atan2}\left(2x_{k,i,j}y_{k,i,j}, x_{k,i,j}^2 - y_{k,i,j}^2\right)}{t_{k,i,j}}, \quad k=1,2 \\
    & v_{k,i,j} = r_j\omega_{k,i,j}, \quad k = 1,2 \\
\end{split}
\end{equation}

\hh{The computation of $\boldsymbol{u}_{k,i,j}=(\omega_{k,i,j}, v_{k,i,j})$ for $\mathcal{T}$ and all the obstacles sets the limits of the DOV. The union of the DOV and the free robot velocity space is the DOVS, represented in Figure~\ref{fig:workspace}.}


\hh{\begin{remark}
The DOVS is used for representing information of the scenario accessible by the robot in $\mathcal{V}$. This representation is therefore valid even though the future trajectories of obstacles are unknown, and the obstacles change their velocities in time. The DOVS is recomputed every control period with the updated information, which is then used for the decision making.
\end{remark}}

The differential-drive velocity constrains of Equations \ref{eq:dymanic-const} and \ref{eq:acc-const} are directly constructed in the DOVS, as shown in Figure \ref{fig:DOVS-rhombus}. This combination is used to learn how to navigate safely, as explained in the following sections.

\subsection{Reinforcement learning setup}\label{sec:DRL-setup}

The problem of robot autonomous navigation in dynamic scenarios may be considered as a sequential decision-making problem to find an efficient policy that leads the robot to the goal while reactively adapting to the changes in the environment to avoid collisions. It may be modeled as a Partially Observable Markov Decision Process (POMDP) with a tuple $(\mathcal{S}, \mathcal{A}, \mathcal{O}, T, O, R)$, where $\mathcal{S}$ is the state space, $\mathcal{A}$ the action space, $\mathcal{O}$ the observation space, $T(\mathbf{s}_{t+1} \mid \mathbf{a}_t, \mathbf{s}_t)$ the transition model, $O(\mathbf{o}_{t+1} \mid \mathbf{s}_{t+1}, \mathbf{a}_t)$ the observation model and $R(\mathbf{s}_t, \mathbf{a}_t)$ the reward function. At every time step $t$, the robot takes an action $\mathbf{a}_t \in \mathcal{A}$ in an environment whose state is $\mathbf{s}_t \in \mathcal{S}$ by following a policy $\pi(\mathbf{s}_t)$. The agent uses information observed from the environment, $\mathbf{o}_t \in \mathcal{O}$, and the action taken modifies the environment given the transition model, resulting the next state $\mathbf{s}_{t+1} \in \mathcal{S}$. The goal of the learning process is estimating the optimal policy $\pi^*$ that maximizes the expected cumulative reward:

\begin{equation}
\begin{split}
 \pi^*(\mathbf{s}_t) = & \arg \max_{\mathbf{a}_t \in A} \left[ R(\mathbf{s}_t, \mathbf{a}_t) + \gamma \sum_{\mathbf{s}_{t+1} \in S} T(\mathbf{s}_{t+1} \mid \mathbf{a}_t, \mathbf{s}_t) \right.  \\
& \left. \sum_{\mathbf{o}_{t+1} \in \mathcal{O}} O(\mathbf{o}_{t+1} \mid \mathbf{s}_{t+1}, \mathbf{a}_t) V(\mathbf{s}_{t+1}) \right] \\
\end{split}
\end{equation}
\noindent where $\gamma$ is a discount factor to balance the present value of future rewards, and $V(\mathbf{s}_{t})$ is the value function, which represents the expected cumulative reward starting from state $\mathbf{s}_t$ and following the policy $\pi(\mathbf{s}_t)$:
\begin{equation}
\begin{split}
 V(\mathbf{s}_t) = & R(\mathbf{s}_t, \pi(\mathbf{s}_t)) + \gamma  \sum_{\mathbf{s}_{t+1} \in S} T(\mathbf{s}_{t+1} \mid \pi(\mathbf{s}_t), \mathbf{s}_t) \\
& \sum_{\mathbf{o}_{t+1} \in \mathcal{O}} O(\mathbf{o}_{t+1} \mid \mathbf{s}_{t+1}, \pi(\mathbf{s}_t))  V(\mathbf{s}_{t+1}) \\
\end{split}
\end{equation}

\subsubsection{State space} 

The state $\mathbf{s}_t \in \mathcal{S}$ is the unobservable true condition of the system that drives the dynamics of the environment. In our problem, we consider that there is a robot with radius $r_{robot}$ position $\mathbf{p}_t$ and velocity $\mathbf{u}_t$ at time $t$ that drives towards a goal $\mathbf{g}$. The rest of the state is the set of points occupied by the set of static and dynamic obstacles $\mathscr{O}_{\forall,t}$, the radius $\mathbf{r}_{obs} \in \mathbb{R}^M$ and velocity $\mathscr{V}_t = \{(v_i, \omega_i) \forall i \in [0 \dots M]\}$ of the $M$ obstacles that are dynamic:

\begin{equation}
    \mathbf{s}_t = \{r_{robot}, \mathbf{p}_t, \mathbf{u}_t, \mathbf{g}, \mathscr{O}_{\forall,t}, \mathbf{r}_{obs}, \mathscr{V}_t\}
\end{equation}

\subsubsection{Action space} 

The action $\mathbf{a}_t \in \mathcal{A}$ should account for the kinematics and dynamics of the robot stated in Equation~\ref{eq:dymanic-const} and Equation~\ref{eq:acc-const}. To do so, we geometrically design an action space that ensures the constraints are not violated by construction. This construction is represented in Figure~\ref{fig:action_space}. Having that the robot has the velocity $\boldsymbol{u}_t$ in the current control period, $\boldsymbol{u}_{t+1}$ must not be outside the green rhombus (the dynamic window of Equation~\ref{eq:acc-const}) and must not surpass $l_1$ nor $l_2$ (Equation~\ref{eq:dymanic-const}). The idea of the action space is solving this problem by selecting velocities that respect the restrictions by definition. First, we construct an action space that make velocities be always in the dynamic window. Second, we constraint the dynamic window if it intersects with constraints in Equation~\ref{eq:dymanic-const}.

 We define a continuous action space of two values $\mathbf{a}_t = (a_{1,t}, a_{2,t})\in [0,1]^2$ that linearly combine $\mathbf{b}_{1,t}$ and $\mathbf{b}_{2,t}$, to compute velocities that always remain inside the dynamic window:

\begin{figure}
    \centering
    \includegraphics[width=\linewidth]{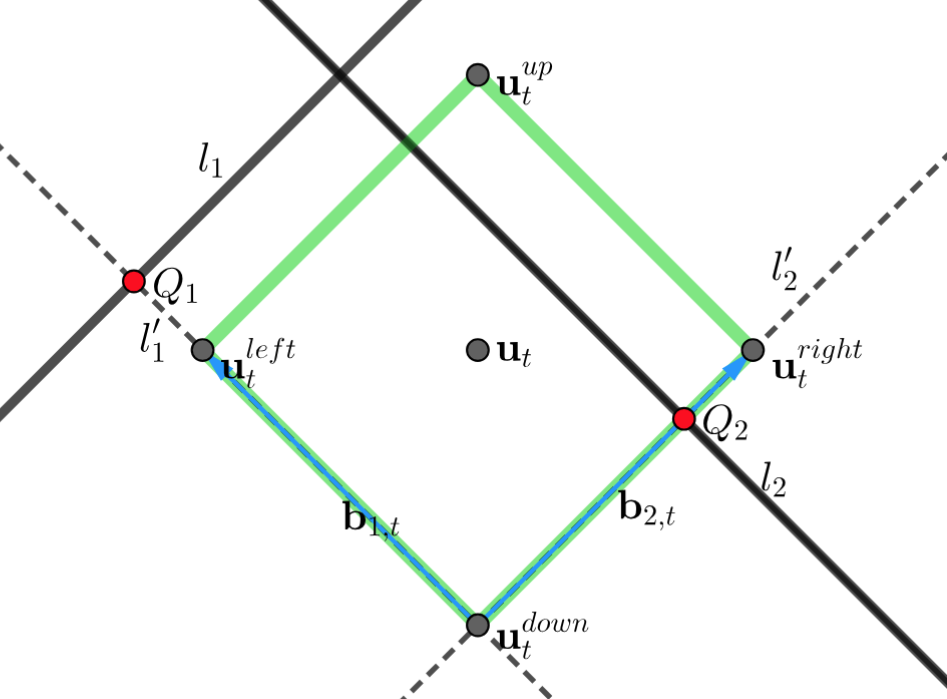}
    \caption{Graphical representation of the notation used in the action space configuration, attaining for restrictions previously represented in Figure~\ref{fig:DOVS-rhombus}. $\mathbf{u}_t$, $\mathbf{u}_t^{up}$, $\mathbf{u}_t^{down}$, $\mathbf{u}_t^{left}$ and $\mathbf{u}_t^{right}$ are represented with gray points, $l_1$ and $l_2$ with black lines, $l_1'$ and $l_2'$ with dotted lines, $\mathbf{b}_{1,t}$ and $\mathbf{b}_{2,t}$ with blue arrows, and $Q_1$ and $Q_2$ with red points.}
    \label{fig:action_space}
\end{figure} 

\begin{equation}
   \mathbf{u}_{t+1} = \mathbf{u}_{t}^{down} + a_{1,t}\mathbf{b}_{1,t} + a_{2,t}\mathbf{b}_{2,t}    
\label{eq:linear-comb}
\end{equation}

To do so, the four corners of the dynamic window are calculated:
\begin{equation}
\begin{split}
   & \mathbf{u}_t^{up} = (\omega_t, {v}_t + a_{max}\Delta t) \\
   & \mathbf{u}_t^{down} = (\omega_t, {v}_t - a_{max}\Delta t) \\
   & \mathbf{u}_t^{left} = (\omega_t - \frac{\omega_{max}a_{max}\Delta t}{v_{max}}, {v}_t) \\
   & \mathbf{u}_t^{right} = (\omega_t + \frac{\omega_{max}a_{max}\Delta t}{v_{max}}, {v}_t) \\
\end{split}
\end{equation}

Using them, the two vectors to form a basis in the velocity space are defined:

\begin{equation}
\begin{split}
   & \mathbf{b}_{1,t} = \mathbf{u}_{t}^{left} - \mathbf{u}_{t}^{down} \\
   & \mathbf{b}_{2,t} = \mathbf{u}_{t}^{right} - \mathbf{u}_{t}^{down} \\
\end{split}
\end{equation}

The kynematic restriction of Equation~\ref{eq:dymanic-const} ($l_1$ and $l_2$ in Figure~\ref{fig:action_space}) are considered by limiting $\boldsymbol{a}_{t}$ maximum value: $a_{1,t} \in [0, a_{1,t}^{max}]$ and $a_{2,t} \in [0, a_{2,t}^{max}]$. This limit is only needed if the dynamic window intersects with $l_1$ or $l_2$. In the velocity space, the lines that pass through $\mathbf{u}_{t}^{down}=(\omega_{t}^{down}, v_{t}^{down})$ and have the direction of $\mathbf{b}_{1,t}$ and $\mathbf{b}_{2,t}$ are $l_1'$ and $l_2'$, respectively:

\begin{equation}
\begin{split}
   & l_1': v = -\frac{{v}_{max}\omega}{{\omega}_{max}} + \frac{v_{max}\omega_t^{down}}{\omega_{max}} + v_t^{down} \\
   & l_2': v = \frac{{v}_{max}\omega}{{\omega}_{max}} - \frac{v_{max}\omega_t^{down}}{\omega_{max}} + v_t^{down} \\
\end{split}
\end{equation}

And the intersection points between $l_1$ and $l_1'$, and $l_2$ and $l_2'$ are $Q_1(\omega_{1,q}, v_{1,q})$ and $Q_2(\omega_{2,q}, v_{2,q})$, with:
\begin{equation}
\begin{split}
   &\omega_{1,q} = \frac{1}{2} (\frac{\omega_{max}v_t^{down}}{v_{max}} + \omega_t^{down} - \omega_{max} ) \\
   & v_{1,q} = \frac{1}{2} (\frac{v_{max}\omega_t^{down}}{\omega_{max}} + v_t^{down} - v_{max} ) + v_{max}\\
   & \omega_{2,q} = \frac{1}{2} (- \frac{\omega_{max}v_t^{down}}{v_{max}} + \omega_t^{down} + \omega_{max} ) \\
   & v_{2,q} = \frac{1}{2} (-\frac{v_{max}\omega_t^{down}}{\omega_{max}} + v_t^{down} - v_{max} ) + v_{max}\\
\end{split}
\end{equation}

The action $\mathbf{a}_t$ will be bounded only if the distance between $\mathbf{u}_{t}^{down}$ and $Q_1$ and $Q_2$ is smaller than $\mathbf{b}_{1,t}$ or $\mathbf{b}_{2,t}$, since the dynamic window will be in collision with $l_1$ or $l_2$. Therefore:
\begin{equation}
\begin{split}
    & a_{1,t}^{max} = \min\left(1, \frac{||P_1 - \mathbf{u}_{t}^{down}||}{||\mathbf{b}_{1,t}||}\right)\\
    & a_{2,t}^{max} = \min\left(1, \frac{||P_2 - \mathbf{u}_{t}^{down}||}{||\mathbf{b}_{2,t}||}\right)\\
    \label{eq:action-bounds}
\end{split}
\end{equation}

\subsubsection{Observation space, observation model and transition model} \label{sec:transition-model}

The observation $\mathbf{o}_t \in \mathcal{O}$ states what the robot perceives from the environment. The robot estimates from the sensor data the values of the state, but its goal and radius, which are already known:
\begin{equation}
    \mathbf{o}_t = \{r_{robot}, \hat{\mathbf{p}}_t, \hat{\mathbf{u}}_t, \mathbf{g}, \hat{\mathscr{O}}_{\forall,t}, \hat{\mathbf{r}}_{obs}, \hat{\mathscr{V}}_t\},
\end{equation}
\noindent where $\hat{\mathbf{p}}_t$ is the estimated robot position, $\hat{\mathbf{u}}_t$ the estimated robot velocity, $\hat{\mathscr{O}}_{\forall,t}$ the estimated position of obstacles, and $\hat{\mathbf{r}}_{obs}$ and $\hat{\mathscr{V}}_t$ the estimated radius and velocity of the moving obstacles.

The robot is modeled to have a 2-D LiDAR laser sensor. It estimates $\hat{\mathbf{p}}_t$ and $\hat{\mathbf{u}}_t$ with the laser data and the odometry, using Adaptive Monte Carlo Localization (AMCL) \cite{fox1999markov}. An extended version of the work proposed by \cite{przybyla2017detection} is used to detect static and dynamic obstacles and estimate $\hat{\mathscr{O}}_{\forall,t}$, $\hat{\mathbf{r}}_{obs}$ and $\hat{\mathscr{V}}_t$ from the 2-D LiDAR measurements. It detects circular and linear obstacles from the scans, by using a grouping and splitting algorithm, associating obstacles of any shape to circles or a set of segments. The circular obstacles are tracked in time with a Kalman filter, wich also estimates its velocity assuming constant velocity in the update step of the Kalman filter. We refer the readers to \cite{przybyla2017detection} for more details about the tracking. This kind of conditions makes the robot face simulated occlusions and estimation errors. 

We assume that the robot follows the deterministic transition model of Equation~\ref{eq:kinematics}, subject to the restrictions stated in Equation~\ref{eq:dymanic-const} and Equation~\ref{eq:acc-const}. We model the dynamic obstacles using circular motion with constant random linear and angular velocity. The obstacles avoid each other using ORCA \cite{van2011reciprocal}, resulting in motion that is different from the constant velocities previously defined. The robot is invisible for the obstacles, making the scenario more challenging. Otherwise, the robot could not learn to avoid non-cooperative obstacles.

\subsubsection{Reward function}

The goal of the agent is reaching the goal while avoiding collisions in the shortest time possible. To achieve this behavior, the reward functions proposed in \cite{long2018towards} and \cite{sathyamoorthy2020densecavoid} are used as inspiration. It is defined with a simple equation that discriminates between terminal and non-terminal states at time $t$:

\begin{equation}
   R(\mathbf{s}_t, \mathbf{a}_t) =  \left\{
    \begin{array}{ll}
        r_{goal}, & d_{t,goal} < 0.15 \\
        r_{collision}, & \textnormal{collision detected} \\
        -r_{dist}\Delta{d_{t,goal}} + r_{t,safedist}, & \textnormal{otherwise}
    \end{array}
    \right. ,
    \label{eq:reward}
\end{equation}
\noindent where $d_{t,goal} = ||\mathbf{g}-(x_t, y_t)||$ and $\Delta{d_{t,goal}} = d_{t,goal} - d_{t-1,goal}$. The robot receives a reward of $r_{goal} = 15$, when it reaches the goal within a certain threshold; and a negative reward $r_{collision}=-15$ when it collides. Reward shaping is used to accelerate training in non-terminal states by encouraging the agent to get closer to the goal ($-r_{dist}\Delta{d_{t,goal}}$ with $r_{dist}=2.5$), and by penalizing the agent if it is too close to an obstacle with:

\begin{equation}
    r_{t,safedist} =  \left\{
    \begin{array}{ll}
        -0.1\lvert0.2-d_{t,obs}\rvert, & d_{t,obs} < 0.2\\
        0, & \textnormal{otherwise}
    \end{array}
    \right. ,
    \label{eq:safedist}
\end{equation}
\noindent where $d_{t,obs}$ is the distance to the closest obstacle.

\subsection{Network} \label{sec:network}

We propose using a Soft Actor Critic (SAC) \cite{haarnoja2018soft} algorithm to solve the problem, and use the DOVS model as part of the input of the network to extract the dynamism of the environment. The implementation of Stable Baselines 3 \cite{stable-baselines3} is used for the SAC. SAC is an off-policy DRL algorithm that uses an actor to determine the policy and a critic to estimate the state-action values, and introduces entropy maximization to help exploration;  it is known for its stability and sample efficiency.

\begin{figure*}[h]
    \centering
    \includegraphics[width=\textwidth, trim={4.7cm 0.0cm 1.0cm 0.0cm},clip]{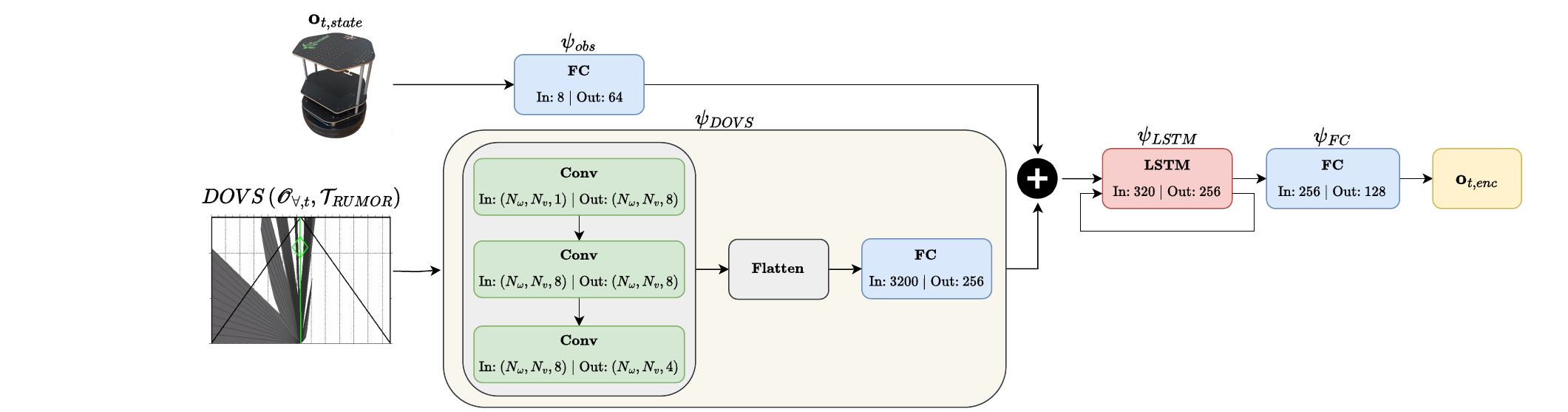}
    \caption{Structure of the encoding network. The DOVS model and the robot state are processed in two different streams, joint later by a memory layer that accounts for previous observations.}
    \label{fig:net}
\end{figure*}

Using raw velocity information of obstacles would require training with obstacles of every possible shape, velocity or radius the robot could face, which is impossible; so using the DOVS to model the scenario improves the adaptation and generalization of the network. The complete structure of the network is represented in Figure~\ref{fig:net}. We process the observation of the environment with two different streams. The first stream uses a discrete set of trajectories formed with velocities ranging from the maximum and minimum velocities of the robot and equally spaced:
\begin{equation}
\begin{split}
    \mathcal{T}_{RUMOR} = & \left\{\tau_j(\omega_j, v_j) \in \mathcal{T} \mid \omega_j \in (\omega_0, \dots, \omega_{N_\omega}), \right.\\
     & v_j \in (v_0, \dots, v_{N_v}), \;  \omega_0 = \omega_{min}, \; \omega_{N_\omega} = \omega_{max} \\
     & \left.  v_0 = v_{min}, \; v_{N_v} = v_{max} \vphantom{\tau_j(\omega_j, v_j) \in \mathcal{T} \mid} \right\} \\
\end{split},
\end{equation}
\noindent where $N_\omega$ is set to 40 and $N_v$ to 20. The DOVS of this set is represented with a grid of $\{-1,1\}$ discrete values indexed by $\omega_j$ and $v_j$ representing whether $\tau(\omega_j, v_j) \in DOV(\mathscr{O}_{\forall, t}, \mathcal{T})$, the unsafe velocities, or $\tau(\omega_j, v_j) \in FV(\mathscr{O}_{\forall, t}, \mathcal{T})$, the safe velocities, respectively. It is processed with a convolutional network of three convolutional layers and a fully connected layer all of them followed by ReLU activation functions, $\psi_{DOVS}$, to preserve the relationships among velocities in the velocity space. Additionally, the following other observation variables $\mathbf{o}_{t,state}$ are separately processed with a fully connected layer with ReLU activation, $\psi_{obs}$:

\begin{equation}
    \mathbf{o}_{t,state}=\left\{\hat{\mathbf{u}}_t, \hat{d}_{t,goal}, \hat{\phi}_{t,goal}, \hat{\mathscr{O}}_{t,dist}, \hat{\mathscr{O}}_{t,\theta}, \hat{\mathscr{O}}_{v,t}, \hat{\mathscr{O}}_{t,dir}\right\},
\end{equation}
\noindent where $\hat{d}_{t,goal}$ and $\hat{\phi}_{t,goal}$ are the distance and angle to the goal with respect to the estimated heading angle of the robot ($\theta_t$), $\hat{\mathscr{O}}_{t,dist}$ the estimated distance of the robot to the closest obstacle, $\hat{\mathscr{O}}_{t,\theta}$ the estimated angle to it, $\hat{\mathscr{O}}_{v,t}$ its estimated velocity and $\hat{\mathscr{O}}_{t,dir}$ its estimated heading angle. Even though the information of these four last variables is already encoded in DOVS, they are included to give more information to the robot in case there is an imminent collision.

\begin{figure}[h]
    \centering
    \includegraphics[width=\linewidth]{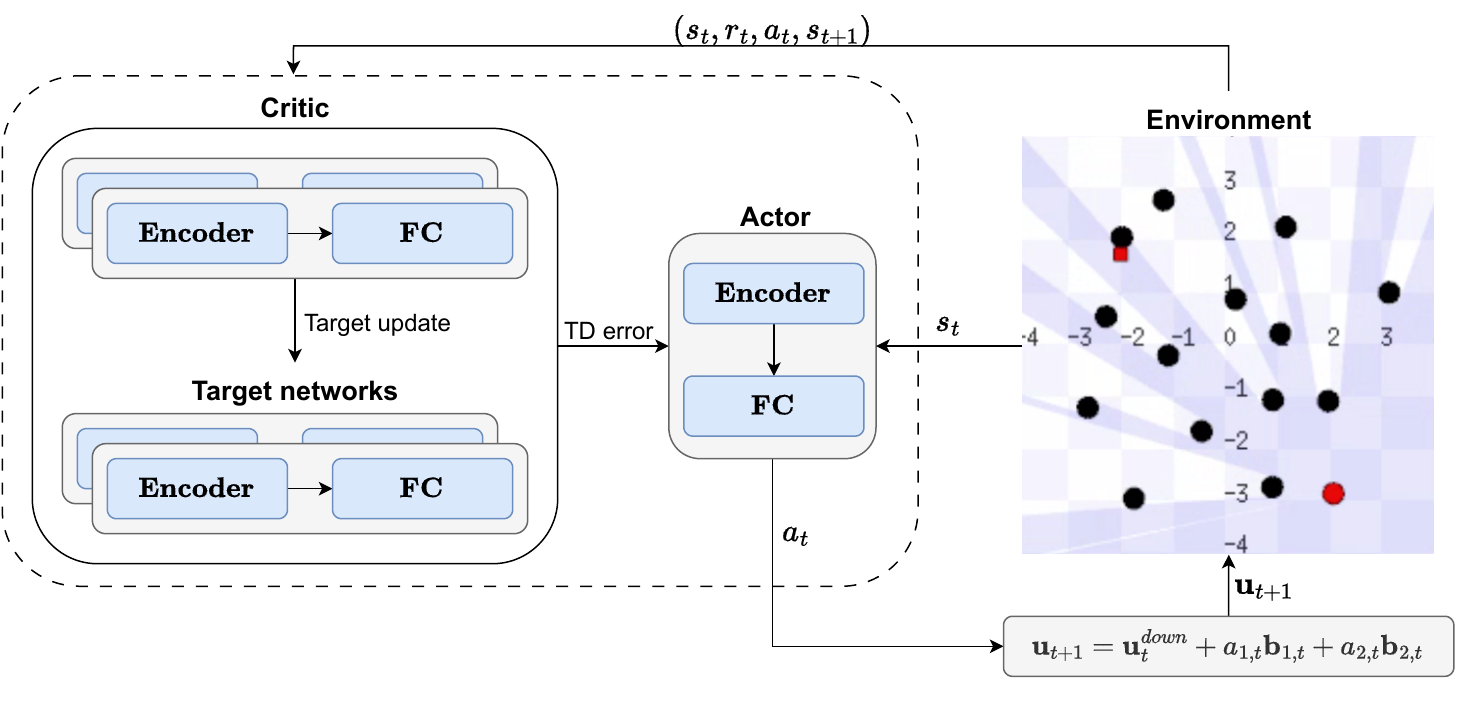}
    \caption{Structure of the SAC architecture used. The critic uses the whole transition for estimating the TD error, used to train the actor. The actor uses the environment observation to select an action, which is translated to a velocity. The environment included is the Stage simulator, used in training.}
    \label{fig:SAC}
\end{figure}

The output of both streams is concatenated and fed to an LSTM layer \cite{hochreiter1997long}, $\psi_{LSTM}$, to save information of previous observation and keep temporal dependencies. The future of the environment is already considered by DOVS, but considering the past could help the network understand changes in the obstacles behavior and make it more robust to estimation errors or occlusions. Finally, a fully connected layer with ReLU activation function, $\psi_{FC}$, produces the final encoded observation $\mathbf{o}_{t,enc}$:
\begin{equation}
\begin{split}
    \mathbf{o}_{t,enc} = \psi_{FC}\left(\psi_{LSTM} \left(\psi_{DOVS}\left(DOVS\left(\mathscr{O}_{\forall, t}, \mathcal{T}_{RUMOR}\right) \right), \right. \right. \\
     \left.\left.\psi_{obs}\left(\mathbf{o}_{t,state}) \right)\right)\right) \\
\end{split}
\end{equation}
An overview of the actor-critic architecture is shown in Figure~\ref{fig:SAC}. It may be seen that the encoder network is duplicated for the actor and the critic, so that they can separately optimize the weights for their objective.

\section{Experiments}\label{sec:results}
\hh{This section details RUMOR evaluation, both with simulation experiments that compare it with existing motion planners (Section~\ref{sec:sim-exp} and hardware experiments that prove its performance in a ground robot (Section~\ref{sec:exp-hardware}.}
\subsection{\hh{Experimental setup}}

\textbf{Simulation setup. }\hh{The planners are tested in a modified version of the Stage simulator \cite{gerkey2003player}. It is also used for training RUMOR and the DRL baselines. This simulator is conveniently integrated in ROS to use AMCL for localization and realistically simulates a LiDAR sensor. The scenario used is an 6x6 m open space with dynamic obstacles the robot has to avoid to reach the goal. The scenario has walls far from this space to let AMCL localize the robot.} The obstacles try to avoid each other using the ORCA algorithm, but cannot see the robot to make it perform the whole obstacle avoidance maneuvers, as stated in Section~\ref{sec:transition-model}. Random positions are sampled for the robot, its goal and the other obstacles. The obstacles have a predefined random linear and angular velocity that try to follow. \hh{The linear velocity of obstacles is sampled between $\frac{v_{max}}{5}$ and $v_{max}$ m/s, where $v_{max}$ is the robot's maximum velocity. This range ensures variability in the obstacle behavior.} The robot velocity limits are set to $v_{max} = 0.7$ m/s and $\omega_{max} = \pi$ rad/s and the maximum acceleration to $a_{max}=0.3$ m/s², similar to those of a Turtlebot 2 platform. \hh{Although typical human walking speeds exceed the obstacles velocity range, setting it to match the robot's maximum speed allows for effective training in obstacle avoidance, despite the TurtleBot 2 being relatively slow. Moreover, the speed of people walking in crowds and when facing other people in open spaces is usually lower than the typical values.}

We \hh{conducted the simulation experiments with} two different perception settings: using the ground truth position of the dynamic obstacles given by the simulator (absolute perception) and using the Kalman Filter-based obstacle detector \cite{przybyla2017detection} (obstacle tracker) previously introduced in Section~\ref{sec:transition-model}, thus facing occlusions and estimation errors. 

\textbf{Training setup.} During training, the robot faces scenarios with increasing number of dynamic obstacles, up to 14, and increasing starting distance to the goal during a first training stage of 1000 episodes. In this way, it learns to reach the goal progressively by using curriculum learning, from simple to more complex tasks. After that stage, it faces 9000 scenarios with a random number of obstacles up to 14. The minimum distance between the robot and the goal is set to 6 m. 

If the simulation takes more than 500 time steps, the episode is finished due to timeout. The episode time step is set to 0.2 s. The network converges in about 10 hours in a computer with a Ryzen 7 5800x processor, a NVIDIA GeForce RTX 3060 graphics card and 64 GB of RAM. The key parameters used are a learning rate of 0.0003 with Adam optimizer \cite{kingma2014adam}, discount factor of 0.99 and soft update coefficient of 0.005. Other parameters can be consulted in the code.

All DRL policies (both RUMOR and those used as baselines) have been trained with this setup. In addition, they have been trained in both perception settings to properly train them, i.e., there are two versions of each of the DRL planners, one trained using absolute perception and the other using the obstacle tracker.

\textbf{Hardware setup. }\hh{RUMOR }was integrated in a Turtlebot 2 platform with a NUC with Intel Core i5-6260U CPU and 8 GB of RAM. The sensor used is a 180º Hokuyo 2D-LIDAR. Due to the real-world design of the system, the same network weights and ROS nodes used in simulation were used in the ground robot, \hh{with the same Kalman Filter-based obstacle detector} and AMCL for localization. \hh{Therefore, the RUMOR version used in the simulation experiments with the obstacle tracker is the same one as the used as in the hardware ones, with the same network weights. Obstacles like walls are detected as segments by the obstacle detector. They are included in the DOVS as colliding velocities by simply computing the velocities that intersect with the segment within a time horizon. This allows the robot to simply detect them without modifying the design.}

\hh{To help the robot navigate along long-paths, through rooms and avoid other non-convex obstacles, we use a way-point generator introduced in \cite{martinez2022full} to feed the robot with intermediate goals rather than simply sending the final goal to it. The idea behind it is first computing a global plan with A$^*$ regarding the static map of the environment and filter the plan to get sparse intermediate points that are corners of sharp turns. A new goal from this set of intermediate goals is sequentially set as the RUMOR goal when the robot is close to approaching the previous one, letting RUMOR compute the motion commands by itself but with intermediate guidance that helps it to avoid convex obstacles. }

\subsection{Simulation experiments}\label{sec:sim-exp}

We conducted quantitative experiments to evaluate the performance of our proposed model. We compared our method, RUMOR, with \hh{DWA~\cite{fox1997dynamic}, TEB~\cite{rosmann2015timed},} ORCA \cite{van2011reciprocal}, \hh{SGAN-MPC~\cite{poddar2023crowd}, }S-DOVS \cite{lorente2018model}, SARL \cite{chen2019crowd} \hh{, RGL~\cite{chen2020relational}, } RE3-RL \cite{10160876} \hh{and SG-D3QN~\cite{zhou2022robot}}. 

\begin{remark}
\hh{The comparison of RUMOR with the baselines planners is unfair, as they do not consider the same velocity constraints. ORCA, SGAN-MPC, SARL, RGL, RE3-RL and SG-D3QN only consider maximum velocity limits, overlooking constraints of Equations 3 and 4. For a more fair comparison, we consider a different version of RUMOR with no restrictions (NR-RUMOR). NR-RUMOR has a continuous action space that comprises the whole robot velocity range: $\mathbf{a}_t = (a_{1,t}, a_{2,t})\in [0,v_{max}]\times[-\omega_{max}, \omega_{max}]$.}
\end{remark}

\begin{remark}\hh{
    We used ROS navigation stack implementations of DWA and TEB, which considers acceleration restrictions in $v$ and $w$ separately. As shown before, the maximum linear acceleration/velocity of a differential-drive robot is directly related to its angular acceleration/velocity. RUMOR accounts for these restrictions, but DWA and TEB do not. The result of this difference may be understood with Figure~\ref{fig:DOVS-rhombus}. First, the restrictions implemented in DWA and TEB do not consider kinematic constraints of Equation~\ref{eq:dymanic-const}, so their velocities may surpass the two black lines. Second, the acceleration constraints would be equivalent to, instead of using the rhombus of Equation~\ref{eq:acc-const} as the dynamic window around the current robot velocity, using a bounding box that surrounds the rhombus. DWA and TEB therefore have an acceleration and velocity space that doubles the size of RUMOR's, which accounts for every differential-drive constraint.\label{rem:dwa-teb}}
\end{remark}

\begin{remark}\hh{
    We selected as agent-based DRL baselines SARL, RGL, RE3-RL and SG-D3QN. Even though we may adapt RUMOR to their restrictions (NR-RUMOR), these baselines may not be adapted to consider the differential-drive constraints. First, they use a discrete action space, so RUMOR action space is not available for them. Second, it is not clear how to restrict a discrete action space avoiding issues in Remark~\ref{rem:dwa-teb}. Third, modifying those methods to use a continuous action space would imply changing their algorithm, which could alter training time or stability.}
\end{remark}

We tested the planners in the same \hh{5}00 random scenarios with low and high occupancy, with 6 and 12 dynamic obstacles, respectively. The 85\% of the obstacles are set to be dynamic and the same set of scenarios is used for every method. 

We mainly evaluate the planners using the two navigation metrics directly encoded in the reward function, which are the success rate (number of times the robot have reached the goal without collisions or timeouts) and the navigation time. There is a trade-off between both metrics. Trying to achieve a low navigation time leads to risky collision avoidance maneuvers, which leads to lower success rates. \hh{Collisions are sometimes unavoidable, particularly in dense environments or when estimation errors occur. In such cases, the robot may encounter trapping situations from which it cannot escape or face obstacles that move unpredictably toward it. Thus, achieving consistently high success rates in complex scenarios is unrealistic, so these results should be interpreted as a benchmark. In real-world settings, where obstacles typically cooperate in collision avoidance or at least do not actively move toward the robot, the success rate is likely to be higher.}

\begin{figure*}[h]
\centering
\includegraphics[width=0.7\linewidth]{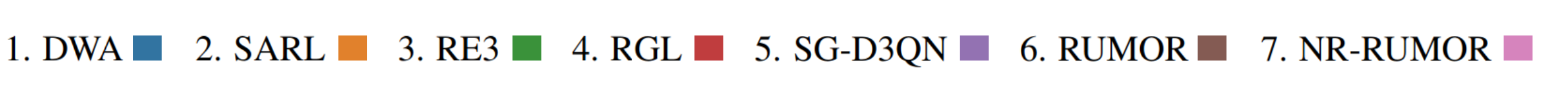} \\
\begin{tabular}{@{}c|c@{}}
    \centering
    \begin{tabular}{@{}cc@{}}
        \footnotesize{Absolute perception (6)} & 
        \footnotesize{Absolute perception (12)}\\
        \hline
        \includegraphics[width=0.23\linewidth]{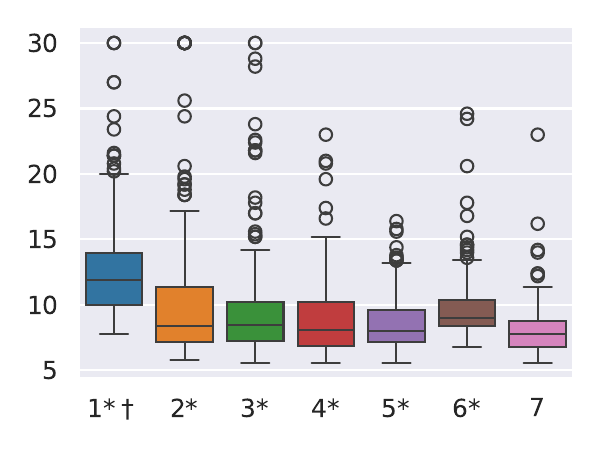} &  
        \includegraphics[width=0.23\linewidth]{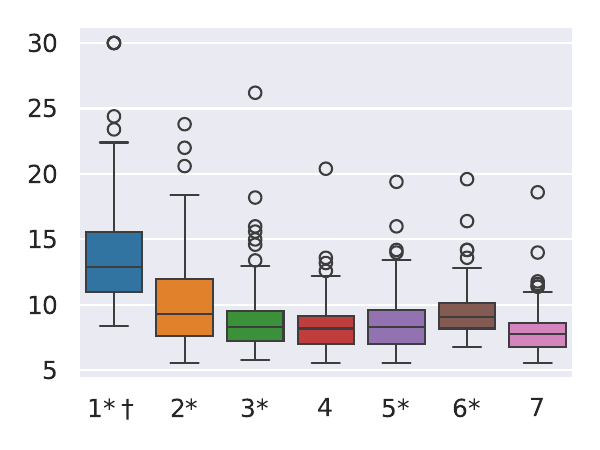} \\
        \footnotesize{(a1) Navigation time} & 
        \footnotesize{(a2) Navigation time}\\

        \includegraphics[width=0.23\linewidth]{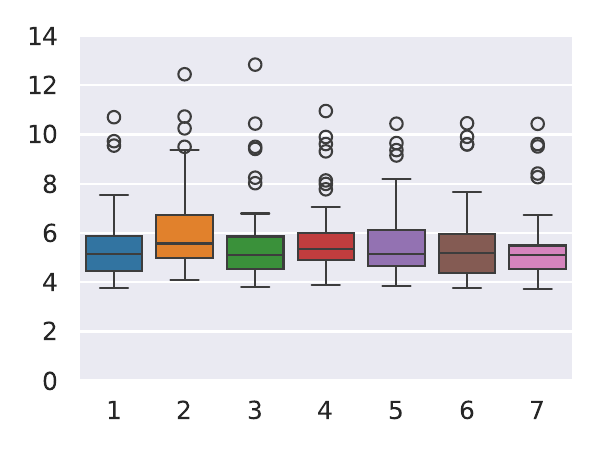} &  
        \includegraphics[width=0.23\linewidth]{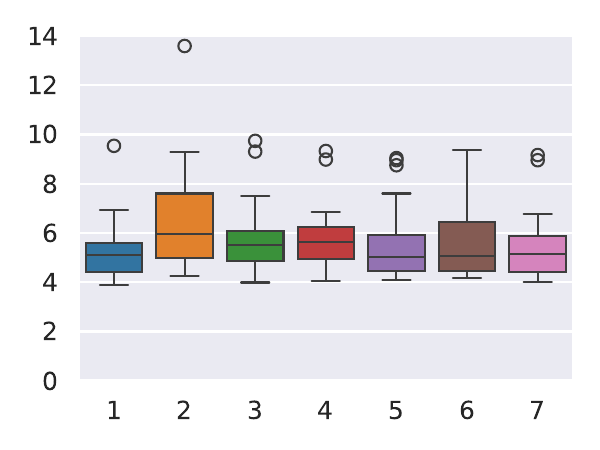} \\
        \footnotesize{(b1) Path length} & 
        \footnotesize{(b2) Path length}\\

        \includegraphics[width=0.23\linewidth]{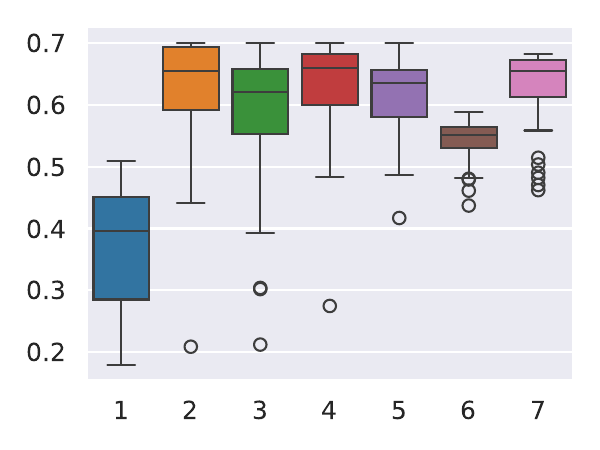} &  
        \includegraphics[width=0.23\linewidth]{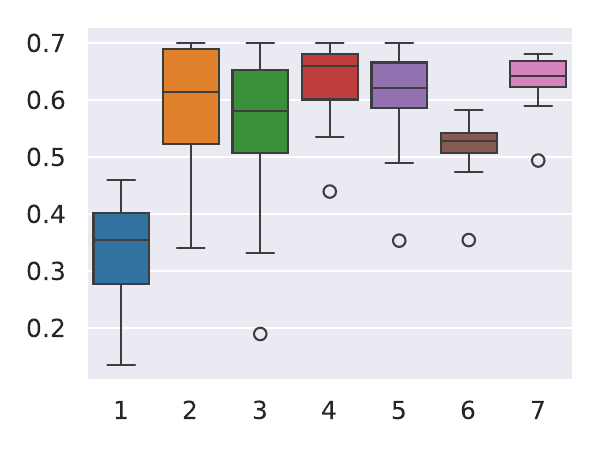} \\
        \footnotesize{(c1) Mean v} & 
        \footnotesize{(c2) Mean v}\\
    \end{tabular}
     &

    \centering
    \begin{tabular}{@{}cc@{}}
        \footnotesize{Obstacle tracker (6)} & 
        \footnotesize{Obstacle tracker (12)}\\
        \hline
        \includegraphics[width=0.23\linewidth]{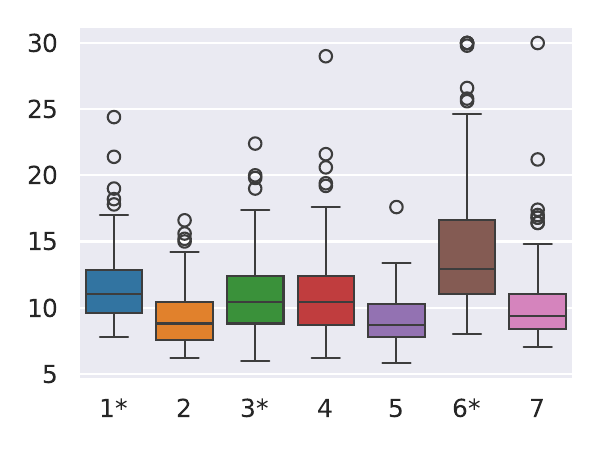} &  
        \includegraphics[width=0.23\linewidth]{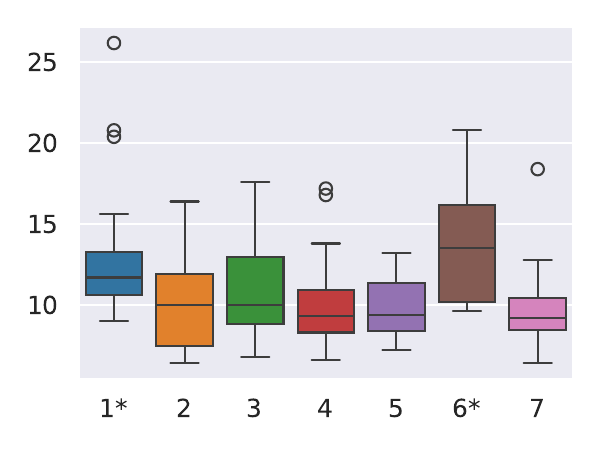} \\
        \footnotesize{(a3) Navigation time} & 
        \footnotesize{(a4) Navigation time}\\

        \includegraphics[width=0.23\linewidth]{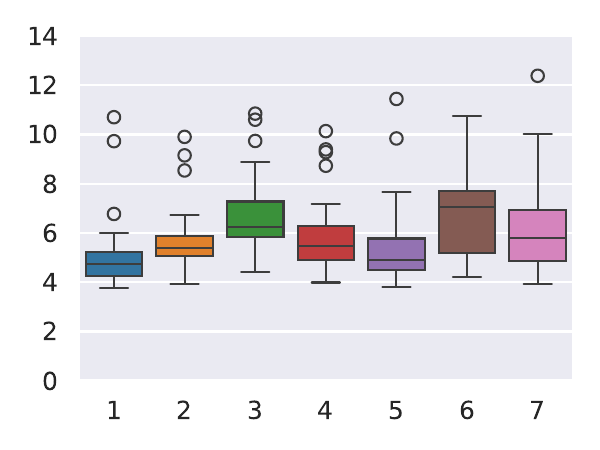} &  
        \includegraphics[width=0.23\linewidth]{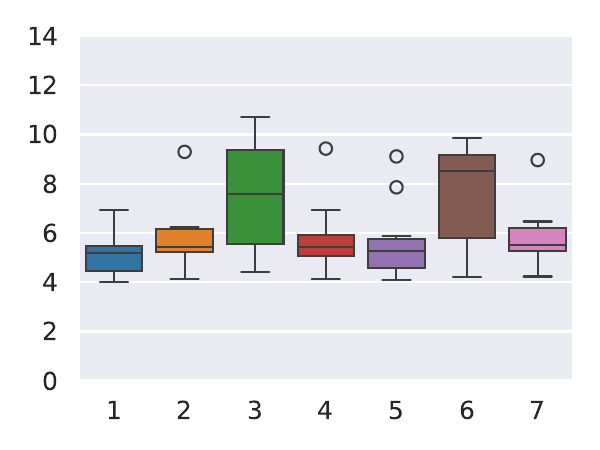} \\
        \footnotesize{(b3) Path length} & 
        \footnotesize{(b4) Path length}\\

        \includegraphics[width=0.23\linewidth]{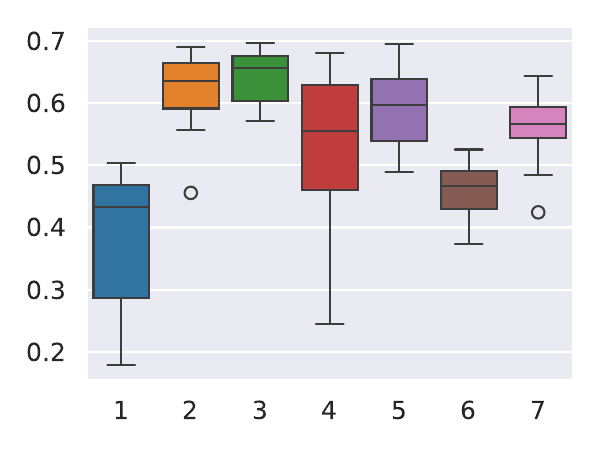} &  
        \includegraphics[width=0.23\linewidth]{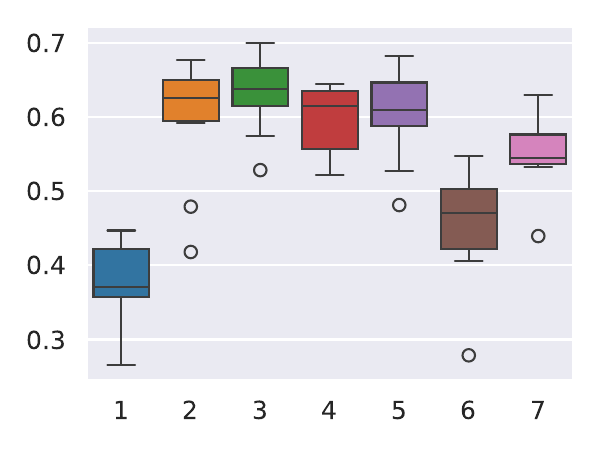} \\
        \footnotesize{(c3) Mean v} & 
        \footnotesize{(c4) Mean v}\\
    \end{tabular} 
    \end{tabular}
    \caption{\hh{Navigation metrics for dynamic scenarios of 6 (left) and 12 (right) obstacles using absolute perception and the obstacle tracker. In (a), $^*$ denotes statistical significant difference between NR-RUMOR and a method, and $^{*\dagger}$ between both NR-RUMOR and RUMOR and a method, with p-value $< 0.05$.}}
    \label{fig:results-12-obs}
\end{figure*}

\hh{The success rates obtained in the experiments are presented in Table~\ref{tab:success_rates}. Scenarios in which all methods failed were excluded from the evaluation metrics.  To assess statistical significance, we performed a $\chi^2$ test to compare the success rate of NR-RUMOR and RUMOR against the baselines. In addition, we recorded the time to reach the goal, mean velocity and path length metrics for each testing episode where all methods succeeded. The results are visualized as box-plots in Figure~\ref{fig:results-12-obs} for DWA and the DRL planners (the rest are not included to simplify visualization). Furthermore, we conducted Mann-Whitney U rank tests with a one-sided ("less") alternative hypothesis to statistically determine if RUMOR and NR-RUMOR achieved significantly lower navigation times compared to the other methods.}

\begin{table}[H]
\centering
\begin{tabular}{c|cc|cc}
 \multirow{2}{*}{\hh{\textbf{Method}}}& \multicolumn{2}{c|}{\hh{\textbf{Abs. Per.}}} & \multicolumn{2}{c}{\hh{\textbf{Obs. Track.}}} \\
 & \hh{\textbf{6 Obs.}} & \hh{\textbf{12 Obs.}} & \hh{\textbf{6 Obs.}} & \hh{\textbf{12 Obs.}} \\
\midrule
\hh{TEB} & \hh{0.47$^{*\dagger}$} & \hh{0.10$^{*\dagger}$} & \hh{0.47$^{*\dagger}$} & \hh{0.10$^{*\dagger}$} \\
\hh{DWA} & \hh{0.69$^{*\dagger}$} & \hh{0.44$^{*\dagger}$} & \hh{0.70$^{*}$} & \hh{0.46$^{*}$} \\
\hh{S-DOVS} & \hh{0.70$^{*\dagger}$} & \hh{0.36$^{*\dagger}$} & \hh{0.58$^{*\dagger}$} & \hh{0.37$^{*\dagger}$} \\
\hh{SGAN-MPC} & \hh{0.66$^{*\dagger}$} & \hh{0.51$^{*\dagger}$} & \hh{0.56$^{*\dagger}$} & \hh{0.36$^{*\dagger}$} \\
\hh{ORCA} & \hh{0.66$^{*\dagger}$} & \hh{0.51$^{*\dagger}$} & \hh{0.60$^{*\dagger}$} & \hh{0.35$^{*\dagger}$} \\
\hh{SARL} & \hh{0.81$^{*\dagger}$} & \hh{0.64$^{*\dagger}$} & \hh{0.74} & \hh{0.46$^{*}$} \\
\hh{RE3} & \hh{0.87$^{*\dagger}$} & \hh{0.71$^{*}$} & \hh{0.64$^{*\dagger}$} & \hh{0.35$^{*\dagger}$} \\
\hh{RGL} & \hh{0.88$^{*}$} & \hh{0.68$^{*}$} & \hh{0.65$^{*}$} & \hh{0.39$^{*\dagger}$} \\
\hh{SG-D3QN} & \hh{0.91} & \hh{0.78} & \hh{0.72$^{*}$} & \hh{0.50$^{*}$} \\
\hh{RUMOR} & \hh{0.91} & \hh{0.72$^{*}$} & \hh{0.70$^{*}$} & \hh{0.46$^{*}$} \\
\hh{NR-RUMOR} & \hh{\textbf{0.94}} & \hh{\textbf{0.82}} & \hh{\textbf{0.78}} & \hh{\textbf{0.58}} \\
\end{tabular}
\caption{\hh{Success rates for the different methods for the simulation experiments, using absolute perception and the obstacle tracker. The method with the best score in each setup is in bold. $^*$ denotes statistical significant difference between NR-RUMOR and a method, and $^{*\dagger}$ between both NR-RUMOR and RUMOR and a method, with a p-value $< 0.05$.}}
\label{tab:success_rates}
\end{table}

\hh{Overall, NR-RUMOR has the highest success rate in all settings, while having lower or at least comparable navigation times. Moreover, the difference with the rest of the planners is statistically significant in most of cases. The only case where the difference is not significant in both low and high density scenarios is when comparing NR-RUMOR with SG-D3QN in absolute perception settings. However, the mean time taken by NR-RUMOR planner to reach the goal is significantly lower in those cases.}

The success rates show that the \hh{agent-}model based planners \hh{(S-DOVS, SGAN-MPC and ORCA)} performance is worse than DRL-based ones \hh{ regarding their success rates}. These planners \hh{use the same agent information DRL planners as input. They} model the environment in every time step, but they are not robust when errors or deep changes in the surroundings occur. 

\hh{Apart from TEB and DWA, which do not use the agent information as input, }every planner shows higher success rates and lower navigation times with absolute perception, due to the absence of partial observability and estimation errors. It is way more complex to predict the future of the environment when facing perception problems. Therefore, their relative performance within the same settings should be compared. It is interesting to see that there is an inconsistency in the performance of SARL and RE3-RL \hh{or RGL} when changing the perception settings. When comparing them using absolute perception, RE3-RL \hh{and RGL} outperform SARL in both success rate and navigation time, while when using the obstacle tracker is the other way around. This is probably due to the difference in the complexity of their networks and the exploration process, degrading the performance of \hh{complex models} in the presence of erroneous observations. In contrast, both NR-RUMOR show a very consistent relative performance no matter the perception settings. \hh{SG-D3QN shows a consistent performance too, but, as stated before, with weaker results than NR-RUMOR.}

RUMOR underperforms NR-RUMOR as expected, as kinematic and dynamic restrictions limits the capacity of maneuvering and sudden reactions. Nonetheless, it displays a better performance, or at least comparable, than the rest of the planners. The difference in the behavior is seen in the velocity and path length graphs. While RUMOR tries to reach the maximum linear velocity as the rest of the planners, the acceleration limits prevents it from achieving it as much time as the rest. In addition, it has to keep a lower velocity in dangerous zones, as it must stop gradually. The path length and navigation time plots show that RUMOR achieve high success rates with longer paths, meaning that it must respect the restrictions to navigate. It follows safer paths where no sudden velocity changes are needed. 

\hh{TEB and DWA performance is the same in both perception settings, as their input, which is the sensor scan and not the individual agent information, is the same in both cases. TEB suffers from the same issues as the other model-based planners. DWA performance is comparable or better than other methods when they suffer from perception errors. It is similar to the restricted RUMOR in those cases. Nevertheless, in absolute perception settings, RUMOR outperforms DWA in both success rate and time to reach the goal with statistical significance, even considering Remark~\ref{rem:dwa-teb}. Thus, it seems like using more accurate perception algorithms (e.g. deep learning) would increase the difference between both methods.}

Two key components of the system explain NR-RUMOR results. First, the input of RUMOR network is the DOVS model of the environment. With 12 surrounding dynamic obstacles, there are unlimited number of possible scenario configurations, and the agent can not be trained to have a robust performance in all of them. On the contrary, DOVS represents the environment in similar terms no matter the scenario, leading to a deeper understanding of the velocity space. When the robot learns to interpret DOVS, it knows how to avoid collisions regardless the amount, positions or velocities of the obstacles. Second, the memory layer of RUMOR network allows it to keep track of previous observations. This is important to overcome partial observability challenges and those derived from sudden changes in the obstacles behavior. Overall, considering every combination in the perception and occupancy settings, NR-RUMOR consistently shows the best performance when compared with the other methods.

\begin{figure}[h]
    \centering
    \begin{tabular}{@{}cc@{}}
        \includegraphics[width=0.47\linewidth]{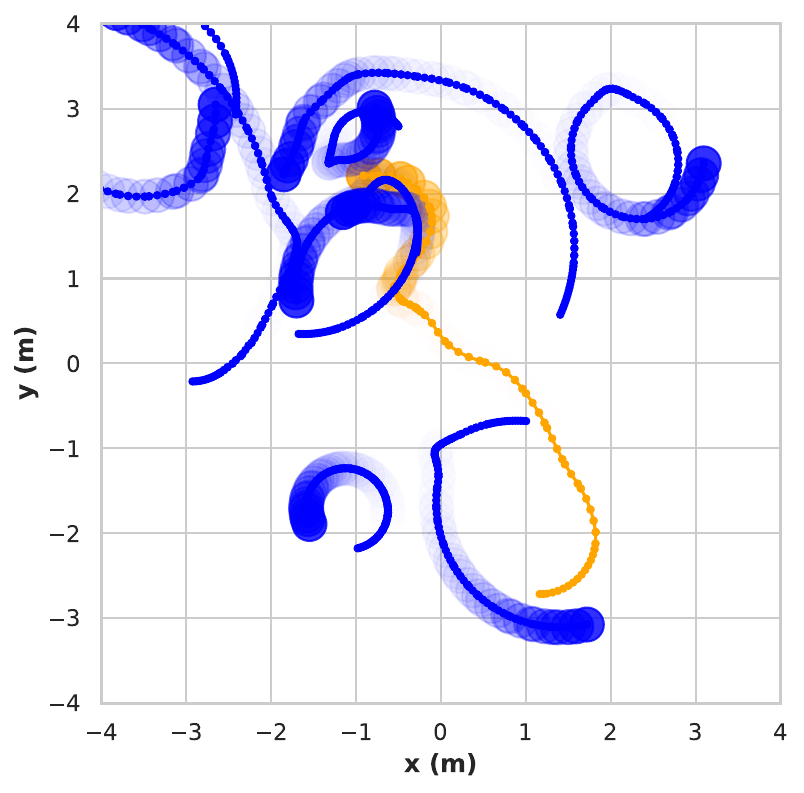} &  
        \includegraphics[width=0.47\linewidth]{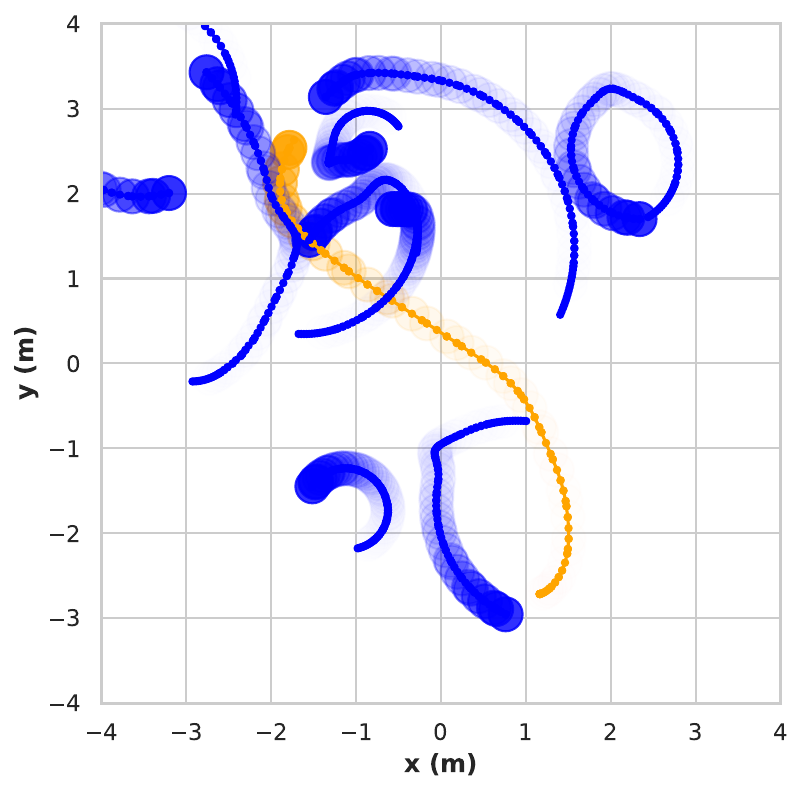} \\
        \footnotesize{(a) NR-RUMOR} & 
        \footnotesize{(b) RUMOR}\\
        
        \includegraphics[width=0.47\linewidth]{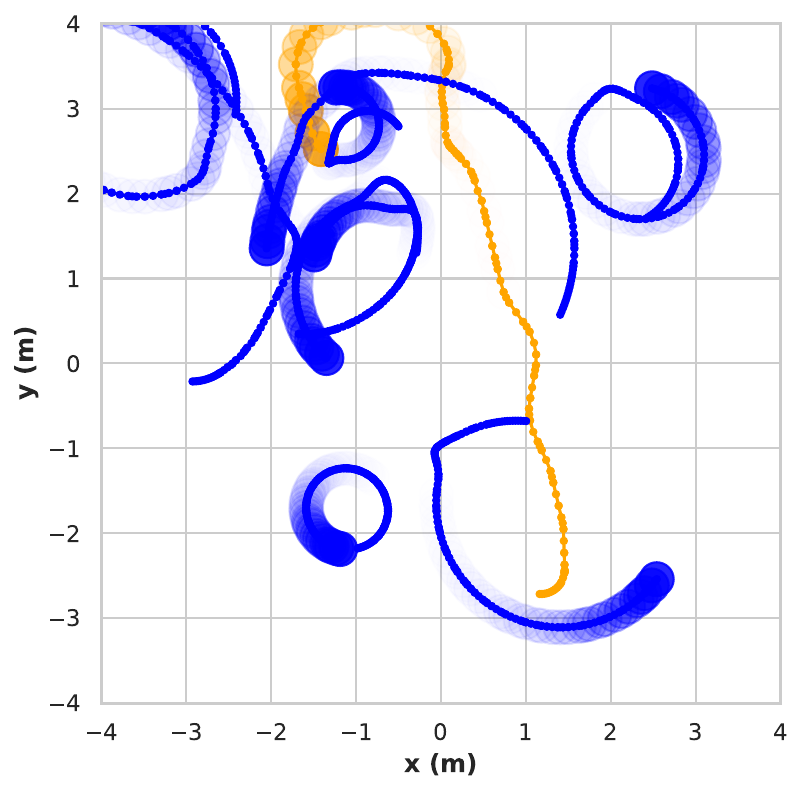} &  
        \includegraphics[width=0.47\linewidth]{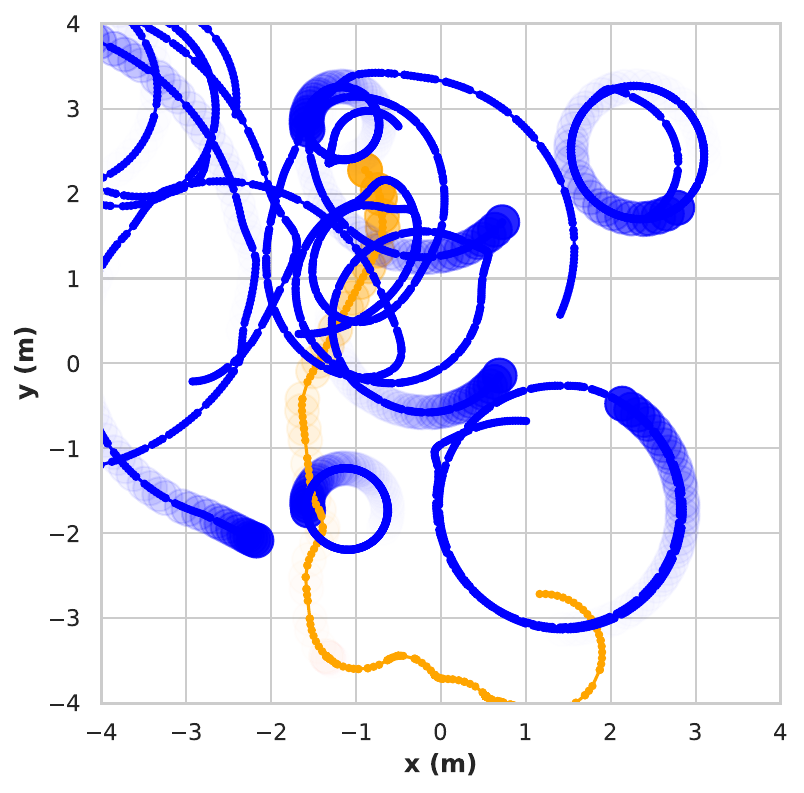} \\
        \footnotesize{(c) RE3-RL} & 
        \footnotesize{(d) SARL}\\
    \end{tabular}
    \caption{A robot (orange) navigating with different planners in a scenario with 9 dynamic obstacles (blue). The evolution in time is represented with increasing opacity, being completely solid at the end of the episode.}
    \label{fig:scenarios}
\end{figure}

An example of the behavior of the different DRL planners is shown in Figure~\ref{fig:scenarios}, in absolute perception settings. All the planners reach the goal successfully, but both RUMOR versions do it in the shortest time, the shortest path lengths and with the most simple trajectories. RUMOR presents the smoothest trajectory, due to the acceleration restrictions, whereas all the other planners present a higher path irregularity. Its action space is the only one that does not allow the robot to reach maximum or minimum velocities at any time step, so it must be consistent with the navigation decisions it makes. In addition, as expected from the metrics results (Figure~\ref{fig:results-12-obs} (b1), (c1), (b3) and (c3)), RE3-RL performs better than SARL.

\subsection{Real-world experiments}\label{sec:exp-hardware}

Hardware experiments were conducted to validate and evaluate RUMOR performance in real conditions. Two different types of real-world experiments were designed, which may be seen in the supplementary video.

\begin{figure}[h]
    \centering
     \begin{tabular}{@{}cc@{}}
         \includegraphics[trim={8.0cm 6.0cm 2.0cm 8.0cm},clip,width=0.47\linewidth]{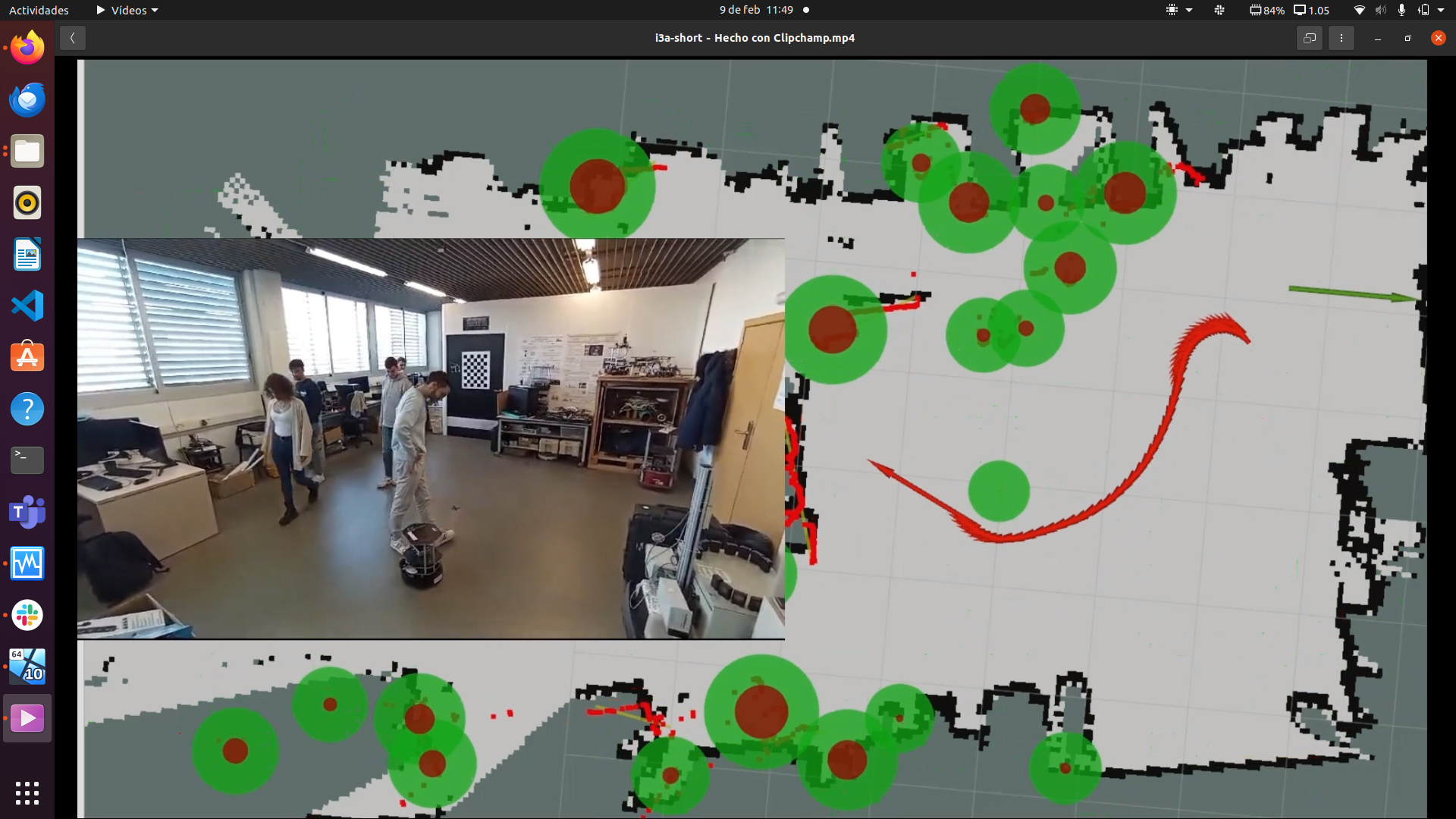}     &     
         \includegraphics[trim={8.0cm 6.0cm 2.0cm 8.0cm},clip,width=0.47\linewidth]{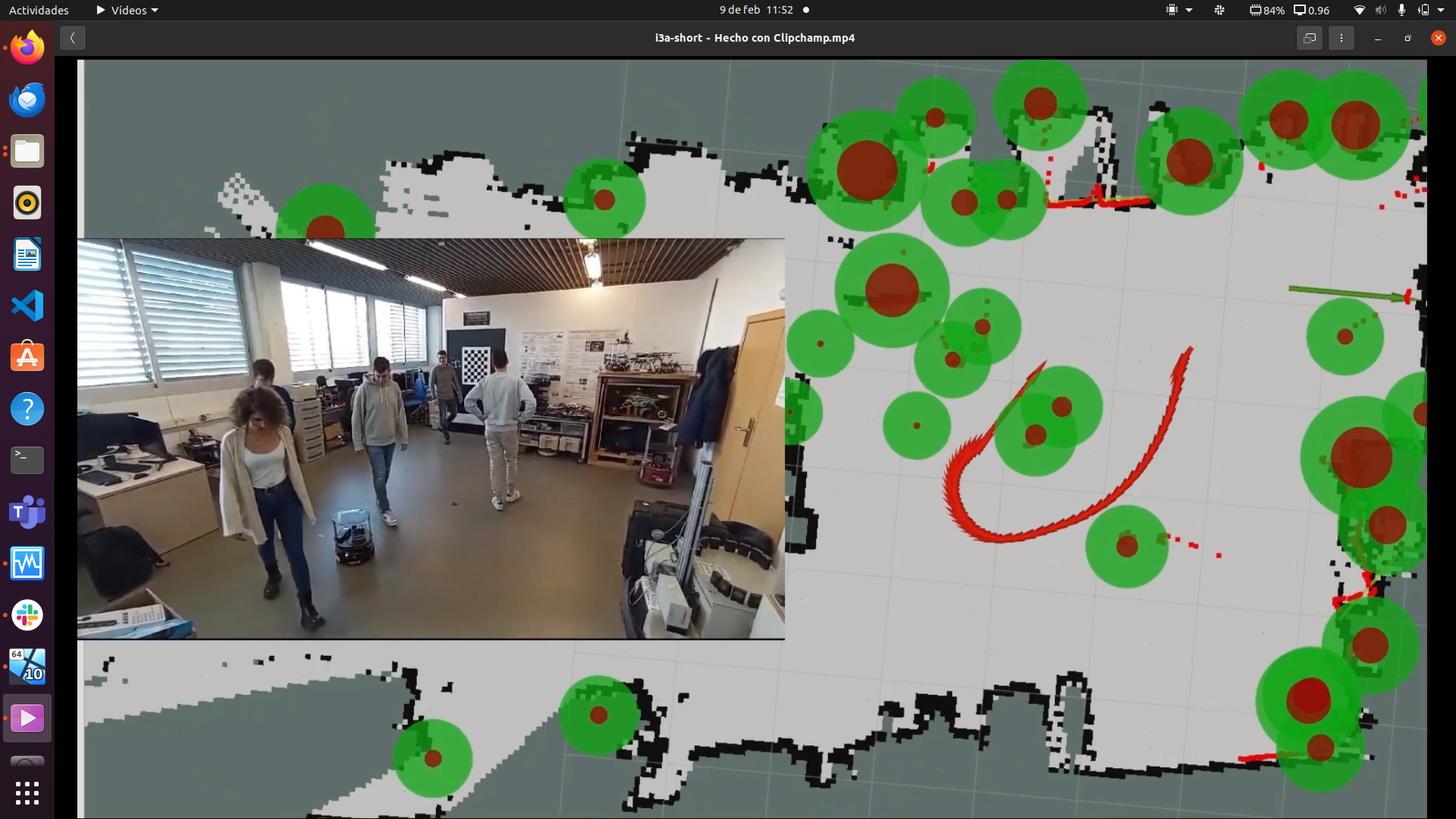}\\
         \footnotesize{(a)} &
         \footnotesize{(b)} \\
         \includegraphics[trim={8.0cm 6.0cm 2.0cm 8.0cm},clip,width=0.47\linewidth]{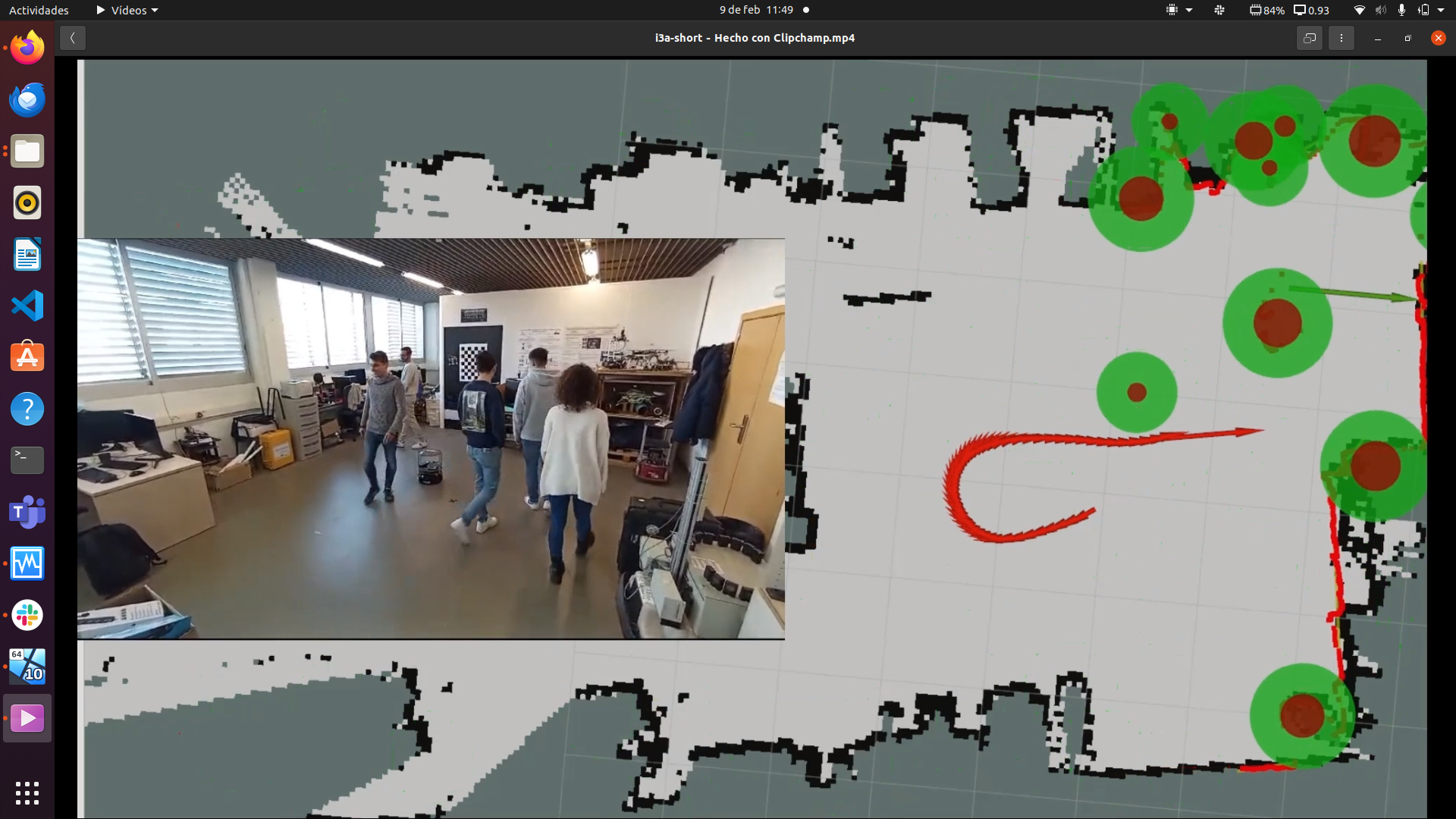}     &     
         \includegraphics[trim={8.0cm 6.0cm 2.0cm 8.0cm},clip,width=0.47\linewidth]{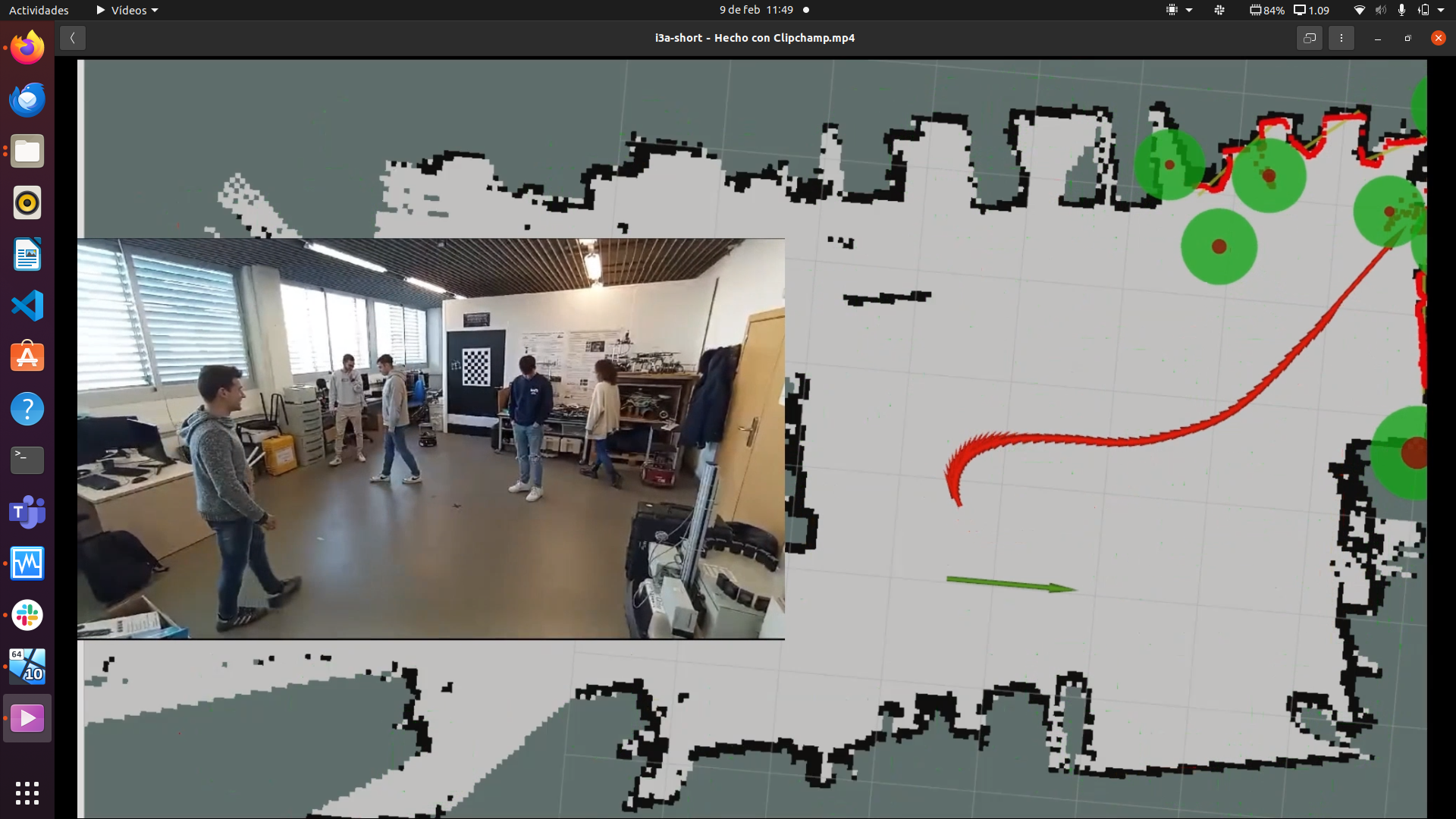}\\
         \footnotesize{(c)} &
         \footnotesize{(d)} \\
     \end{tabular}
        \caption{Photos superposed to the visualization of four consecutive moments of a real-world experiments. The robot position is represented with a red arrow, its previous trajectory with small red arrows, the LiDAR scans with red points, the obstacle positions and radius extracted with the obstacles tracker with red and green circles and the goal of the robot with a green arrow. }
        \label{fig:i3a-exp}
\end{figure}

A set of experiments were carried out to test navigation in very dense dynamic environments. The experimental setup was a scenario with dimensions of about 6x5 m. The robot was tasked to navigate towards dynamically assigned goals in the corners of the room while encountering dynamic obstacles. Pedestrians randomly wandering within the area were used as obstacles. To ensure the validity of the avoidance maneuvers, the individuals were instructed not to visually attend to the robot's movements. Figure~\ref{fig:i3a-exp} shows images of the robot smoothly and safely navigating between two goals. In that specific situation, the robot must reach the goal avoiding collisions with five pedestrians and the limits of the room. The robot first waits for the first person encountered to pass (a). Then, it adapts its trajectory to pass between two people (b) and keeps the collision avoidance maneuver to avoid another one (c). Finally, it continues its way to the goal (d). RUMOR is able to use the DOVS representation of the dynamism of the environment to choose natural and smooth trajectories with low path irregularities to reach the goal, instead of abruptly reacting to obstacles.

\begin{figure}[h]
    \centering

    \begin{tabular}{@{}cc@{}}
        \includegraphics[width=0.47\linewidth]{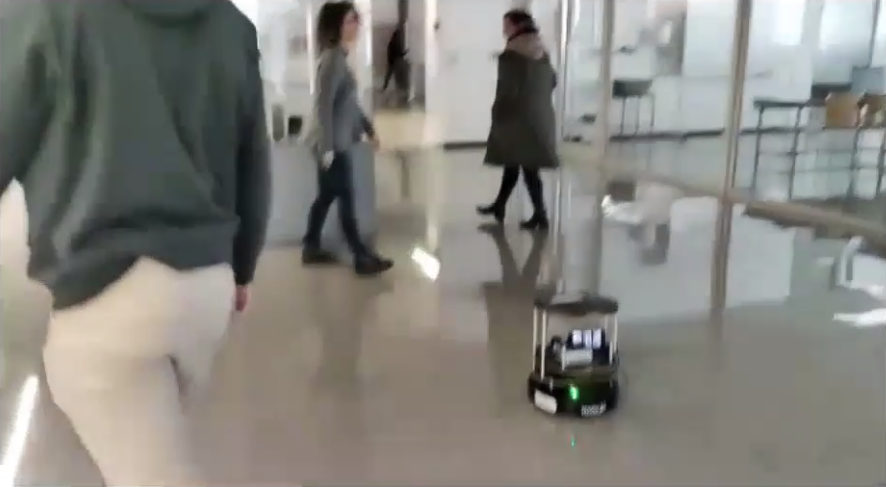} & \includegraphics[width=0.47\linewidth]{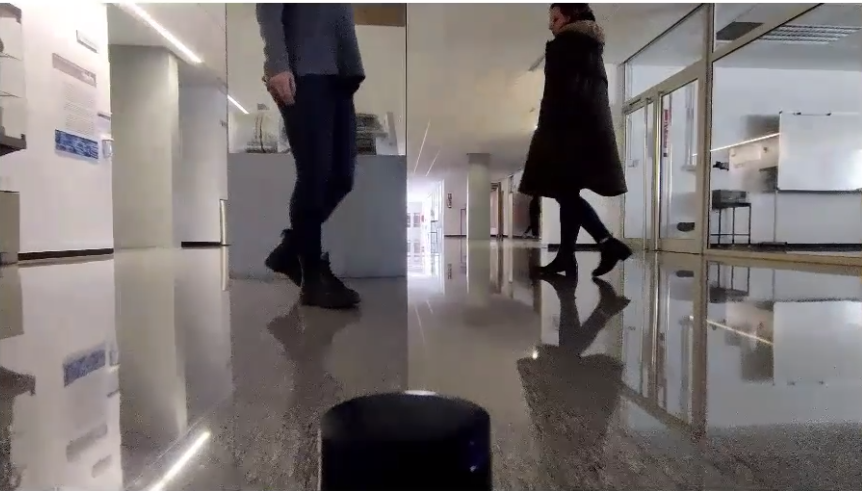} \\
       \footnotesize{(a) Third person point of view}  &  \footnotesize{(b) Robot's point of view}
    \end{tabular}
        \caption{Images taken from the corridor experiment.}
        \label{fig:ada}
\end{figure}

Finally, a setup different from a lab scenario was tried, in settings including a corridor with some rooms and intersections, obstacles with different shapes and people with heterogeneous behaviors, as in real life, to prove that the system works in a completely unseen environment that has not been specifically used in simulation. This may be seen in Figure~\ref{fig:ada}. The results show that the robot is able to navigate very smoothly through the dynamic obstacles that have behaviors different from those seen in training, avoids static obstacles that had not been seen before, and works with the global planner to be able to reach goals that are very far from the initial position. The use of  DOVS, instead of being a complete end-to-end planner, limits out-of-distribution observations, as apparently different scenarios could result in similar DOVS representations.

\section{Conclusion}\label{sec:conclusion}

This work presents a novel motion planner for dynamic environments. It uses the DOVS model to extract the dynamism of the environment, abstracting from specific sensor data directly used by end-to-end approaches, and DRL to select motion commands that efficiently leads the robot to a goal while avoiding collisions with moving obstacles. Instead of directly use the raw obstacle information, which may lead to out-of-distribution observations in scenarios different from the ones seen in training, the system uses DOVS, which encodes it in similar terms regardless the scenario. In addition, we propose a training framework that puts the agent in real-world setting, an advanced DRL algorithm that uses memory to mitigate partial observability issues and a novel action space that intrinsically considers differential-drive kinodynamics, in order to mitigate sim2real transition. The proposed system is tested in random scenarios with  absolute perception (ground truth from the simulator) and partial observability conditions, outperforming existing methods and showing the benefits of combining a model-based approach with a DRL controller. The algorithm is also tested in a ground robot in dense, dynamic and heterogeneous scenarios. 

Future work could include increasing the robustness of the method in scenarios where the obstacle motion is very irregular and unnatural, extending the model to collaborative collision avoidance with multi-robot navigation or adapting the model for 3-D navigation and UAVs. \hh{In addition, we believe that studying the impact of the perception errors in both training and deployment separately would be of great interest; considering a deeper study of different perception algorithms, the effect in neural networks with different complexity, the inclusion of explicit uncertainty estimation and the evolution of the DOVS in presence of noise.}

\section*{CRediT authorship contribution statement}

\textbf{Diego Martinez-Baselga}: Conceptualization, Methodology, Software, Validation, Formal analysis, Investigation, Resources, Data curation, Writing-Original draft preparation, Writing-Reviewing and Editing, Visualization. \textbf{Luis Riazuelo}: Conceptualization, Methodology, Validation, Formal analysis, Investigation, Writing-Reviewing and Editing, Supervision. \textbf{Luis Montano}: Conceptualization, Methodology, Validation, Formal analysis, Investigation, Writing-Reviewing and Editing, Supervision, Project administration, Funding acquisition. 

\section*{Declaration of competing interest}

The authors declare that they have no known competing financial interests or personal relationships that could have appeared to influence the work reported in this paper.

\section*{Data availability}

Data will be made available upon acceptance.

\section*{Acknowladgement}

This work was partially supported by MICIU/AEI/
10.13039/501100011033 and ERDF/EU by grant PID2022-139615OB-I00 and grant PRE2020-094415, and Government of Arag\'{o}n under grant DGA T45-23R.



\bibliographystyle{elsarticle-num} 
\bibliography{references}






\end{document}